\documentclass[10pt,twocolumn,letterpaper]{article}

\usepackage{cvpr}              

\usepackage{booktabs} 
\usepackage{xcolor,colortbl}
\usepackage{amsmath}
\usepackage[accsupp]{axessibility}

\definecolor{cvprblue}{rgb}{0.21,0.49,0.74}
\usepackage[pagebackref,breaklinks,colorlinks,allcolors=cvprblue]{hyperref}

\def\paperID{*****} 
\def\confName{CVPR}
\def\confYear{2025}

\title{Test-time augmentation improves efficiency in conformal prediction}

\author{Divya Shanmugam, Helen Lu, Swami Sankaranarayanan, John Guttag\\
Massachusetts Institute of Technology, CSAIL \\
{\tt\small \{divyas, helenlu, swamiviv, guttag\}@mit.edu}
}

\begin{document}
\maketitle
\begin{abstract}
A conformal classifier produces a set of predicted classes and provides a probabilistic guarantee that the set includes the true class. 
Unfortunately, it is often the case that conformal classifiers produce uninformatively large sets. 
In this work, we show that test-time augmentation (TTA)--a technique that introduces inductive biases during inference--reduces the size of the sets produced by conformal classifiers.  
Our approach is flexible, computationally efficient, and effective. It can be combined with any conformal score, requires no model retraining, and reduces prediction set sizes by 10\%-14\% on average. 
We conduct an evaluation of the approach spanning three datasets, three models, two established conformal scoring methods, different guarantee strengths, and several distribution shifts to show when and why test-time augmentation is a useful addition to the conformal pipeline.
\end{abstract}    

\section{Introduction}


Conformal prediction has emerged as a promising way to equip existing classifiers with statistically valid uncertainty estimates. It does so by replacing the prediction of the most likely class with an \emph{uncertainty set}--a set of classes accompanied by a probabilistic guarantee that the true class appears in the set \citep{shafer_tutorial_2008}. 


Conformal prediction faces two limitations in practice. First, achieving a suitably strong guarantee often leads to prediction sets that are uninformatively large \citep{cresswell2024conformal,zhang2024evaluating}. Large prediction sets have been shown to \emph{decrease} performance when provided as a decision-making aid \citep{babbar_utility_2022,zhang2024evaluating,cresswell2024conformal}.

Second, conformal classifiers inherit the instability of the underlying models. As a result, prediction sets can change significantly in response to small input perturbations, a well-known weakness of neural networks \citep{szegedy2013intriguing}. Applying a horizontal flip to each image in ImageNet, for example, changes the prediction set sizes for 75\% of examples at a coverage guarantee of 99\%. Such behavior also represents a barrier to broader use. 


In this work, we show that test-time augmentation (TTA), a widely-used technique in computer vision, has the potential to address both limitations. TTA involves generating an ensemble of predictions by perturbing the input with label-preserving transformations. It has previously been shown that TTA can be used to make non-conformal classifiers more robust to small input perturbations \citep{cohen2024simple}, more accurate \citep{shanmugam_better_2021}, and better calibrated \citep{hekler_test_2023}. However, previous work has not explored the utility of TTA in the context of conformal prediction.

We propose \emph{test-time-augmented conformal prediction}, which transforms a classifier's predictions using a learned test-time augmentation policy prior to conformal classification. By using distinct sets of labeled data to learn the test-time augmentation policy and the conformal classifier, we preserve the assumption of exchangeability, and thereby the coverage guarantee associated with the conformal predictor.

In experiments testing the performance of conformal predictors subject to distribution shift, we see that test-time augmentation reduces prediction set sizes by 14\% on average, with no loss of coverage. And even when there is no distribution shift, we see a reduction of 10\% on average.
Moreover, we find that classes with the largest average prediction set sizes benefit most from the introduction of test-time augmentation.
We also show that test-time augmentation can bridge gaps between classifiers of different sizes (e.g., we show that test-time augmentation combined with ResNet-50 produces smaller set sizes than ResNet-101 without test-time augmentation).

Our analysis of \emph{why} test-time augmentation reduces prediction set sizes reveals a previously unknown effect of test-time augmentation. Specifically, TTA increases the predicted probability of the true class \emph{even when it is predicted to be unlikely} (for example, promoting the true class from 200th most likely to 100th most likely). Although such behavior has no impact on the maximum predicted probability --- commonly the focus of literature on test-time augmentation \citep{ayhan_test-time_2018,shanmugam_better_2021,perez_enhancing_2021} --- it is valuable in conformal prediction. This is because when the true class is promoted to a higher rank among a classifier's predicted probabilities, the conformal classifier includes fewer incorrect classes to meet the conformal guarantee.

\begin{figure*}
\begin{center}
\includegraphics[width=\textwidth]{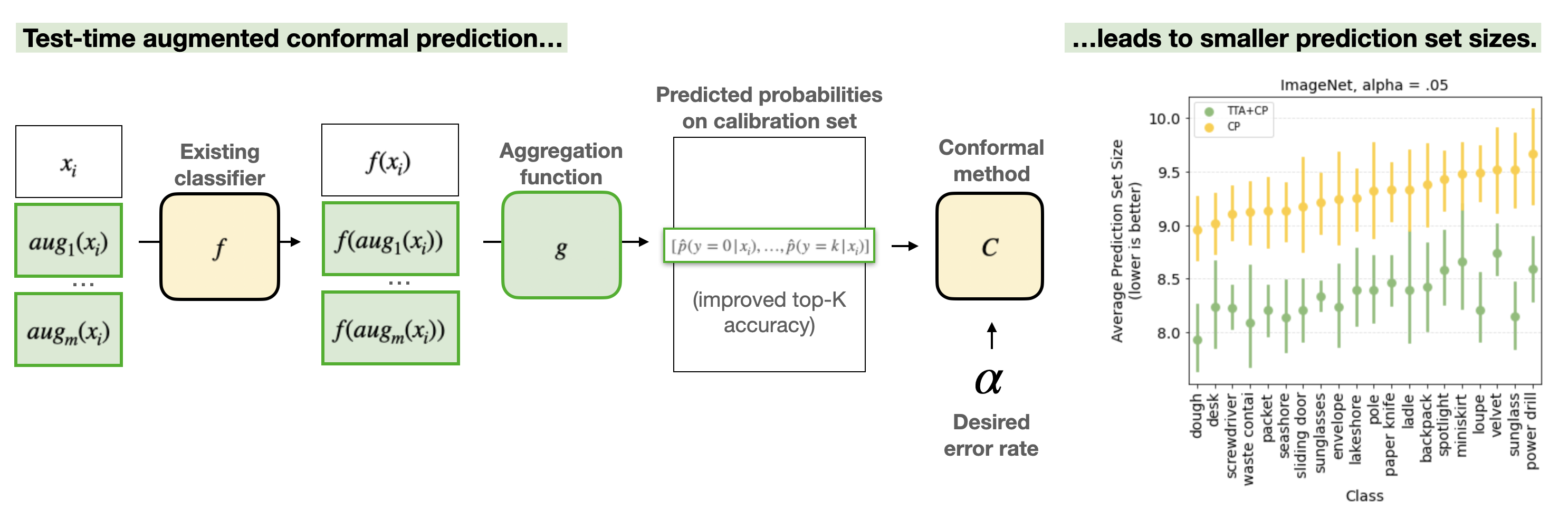}
\end{center}
\caption{We illustrate the addition of test-time augmentation to conformal calibration in green (left) and provide a snapshot of the improvements it can confer (right). We show results on Imagenet, with a desired coverage of 95\%, for the 20 classes with the largest predicted set sizes on average (computed over 10 calibration/test splits).}
\label{fig:intro}
\end{figure*}

\paragraph{Contributions} We make three contributions. To begin, this is the first work to propose combining test-time augmentation, a popular technique in computer vision, with conformal prediction. Second, we present a method that reduces the prediction set sizes of existing conformal predictors by using \emph{automatically learned} test-time augmentations. 
Finally, we demonstrate, in an extensive set of experiments, that our approach to combining conformal prediction and test-time augmentation leads to smaller prediction sets.

\section{Related work}

Conformal prediction was first introduced by \citet{gammerman_learning_1998}, and further developed by \citet{saunders_transduction_1999} and \citet{vladmir_vovk_algorithmic_2005}. We review efforts to ensemble conformal predictors and efforts to reduce prediction set sizes below.

\paragraph{Ensembles in conformal prediction} Several methods that generate ensembles of conformal predictors are known to improve efficiency. These methods include cross-conformal prediction \citep{vovk_cross-conformal_2012}, bootstrap conformal prediction \citep{vovk_cross-conformal_2015}, aggregated conformal prediction \citep{carlsson_aggregated_2014,linusson_calibration_2017}, and out-of-bag conformal prediction \citep{linusson_efficient_2020}. 
The approaches primarily differ in how data is sampled to create the training dataset for the classifier and the calibration dataset for the conformal predictor. 
However, all require training multiple base classifiers or conformal predictors. 
Our approach is distinct: we propose a technique to generate an ensemble from a \emph{single} model by perturbing the input, which requires no additional base models and no additional conformal predictors.

\paragraph{Efficiency in conformal prediction} There are two ways to improve efficiency in split conformal prediction: adjustments to the conformal score or improvements to the underlying classifier. 
Many works have proposed new procedures to estimate  and apply thresholds on conformal scores \citep{tibshirani_conformal_2019,bellotti_optimized_2021,angelopoulos_uncertainty_2022,prinster_JAWS_2022, ding_class-conditional_2023}. \citet{romano_classification_2020} proposed APS, a conformal score based on the cumulative probability  required to include the correct class in a prediction set. \citet{angelopoulos_uncertainty_2022} built on this work to propose RAPS, which modifies APS by penalizing the inclusion of low-probability classes.  
Comparatively little work has focused on improvements to the underlying model. 
\citet{jensen_ensemble_2022} ensemble a set of base classifiers, where the classifiers are created by training models on subsets of the training data. 
\citet{stutz_learning_2022} provide a new way to train the base classifier and conformal predictor jointly through a novel conformal training loss. 
In contrast, our work focuses on improving the underlying model \emph{without} retraining, and can be easily combined with any of the above procedures.

\paragraph{Test-Time Augmentation}

Test-time augmentation (TTA) is a popular technique to improve the accuracy, robustness, and calibration of an existing classifier by aggregating predictions over a set of input transformations \citep{shanmugam_better_2021, perez_enhancing_2021,zhang_memo_2022,enomoto_dynamic_2023,ayhan_test-time_2018,conde_approaching_2023,hekler_test_2023}.
TTA has been applied to a diverse range of predictive tasks across domains ranging from healthcare  \citep{cohen_ICU_2021} to content moderation \citep{lu_improved_2022}. Consequently, many have proposed new ways to perform TTA---for example, learning when to apply TTA \citep{mocerino_adaptive_2021}, which augmentations to use \citep{kim_learning_2020, lyzhov_greedy_2020,chun_cyclic_2022}, and how to aggregate the resulting predictions \citep{shanmugam_better_2021,chun_cyclic_2022,conde_approaching_2023}. Existing work typically focuses on test-time augmentation's impact on highest predicted probability. Here, we analyze how test-time augmentation increases the predicted probability assigned to the true class when it appears \emph{outside} the top few classes, and how that change is consequential in conformal prediction.

\section{Problem setting}
We operate within the split conformal prediction framework. In this setting, a conformal classifier $\mathcal{C}(x_i) \subset \{1, \dots, K\}$ maps input $x_i$ to a subset of $K$ possible classes and requires three inputs:

\begin{itemize}
    \itemsep 0em
    \item \textbf{Calibration set} $D^{(cal)} = \{(x_1, y_1), \dots, (x_n, y_n)\}$, containing $n$ labeled examples.
    \item \textbf{Classifier} $f : \mathcal{X} \mapsto \Delta^K$, mapping input domain $\mathcal{X}$ to a probability distribution over $K$ classes.
    \item \textbf{Desired upper bound on error rate} $\alpha \in [0, 1]$, where $(1 - \alpha)$ is the probability the set contains the true class.
\end{itemize}

We study the introduction of two variables drawn from the test-time augmentation literature:

\begin{itemize}
    \itemsep 0em
    \item \textbf{Augmentation policy} $\mathcal{A} = \{a_0, \dots, a_{m-1}\}$, consisting of $m$ augmentation functions, where $a_0$ is the identity transform. 
    \item \textbf{Aggregation function} $\hat{g}$, which aggregates a set of predictions to produce one prediction.
\end{itemize}

Each variable translates to a key choice in test-time augmentation: what augmentations to apply ($\mathcal{A}$) and how to aggregate the resulting probabilities ($\hat{g}$).

\section{Approach}

  Our goal is to learn an aggregation function $\hat{g}$ to maximize the accuracy of the underlying classifier, and ultimately reduce the sizes of the prediction sets generated from the classifier's predicted probabilities. 
We briefly outline the conformal approach, and then detail the mechanics of our method (illustrated in Figure \ref{fig:intro}). For a detailed introduction to conformal prediction, refer to \citet{shafer_tutorial_2008}. 

 Conformal predictors accept three inputs: a probabilistic classifier $f$, a calibration set $\mathcal{D}^{(cal)}$, and a pre-specified error rate $\alpha$.
Using the these inputs, one can construct a conformal predictor in three steps: 

\begin{enumerate}
    \itemsep 0em
    \item Define a score function $c(x_i,y_i)$, which produces a \emph{conformal score} representing the uncertainty of the input example and label pair. 
    \item Produce a distribution of conformal scores by computing $c(x_i, y_i)$ for all $(x_i, y_i) \in \mathcal{D}^{(cal)}$. 
    \item Compute threshold $\hat{q}$ as the $\lceil (n+1)(1-\alpha) \rceil /  n$ quantile of the distribution of conformal scores over $n$ examples in the calibration set, combined with $\{\infty\}$.
\end{enumerate}

For a new example $x$, we compute $c(x_i, y)$ for all $y \in \{1, \dots, K\}$, and include all $y$ for which $c(x_i,y) < \hat{q}$.  We adopt the conformal score proposed by \citet{romano_classification_2020}, which equates to the cumulative probability required to include the correct class:

\begin{align}
    \hat{\pi}_x(y') &= \hat{p}(y = y' | x) = f(x)_{y'} \label{eq:1} \\
    \rho_x(y) &= \sum_{y' = 1}^K \hat{\pi}_x(y') \mathbb{I}[\hat{\pi}_x(y') > \hat{\pi}_x(y)] \\
    c(x, y) &= \rho_x(y) + u \cdot \hat{\pi}_x(y) 
\end{align}
\label{eqn:c_score}

where $\rho_x(y)$ is the cumulative probability of all classes predicted with higher probability than $y$ and $ \hat{\pi}_x(y')$ corresponds to the predicted probability of class $y'$ given $x$. The variable $u \sim U(0,1)$. Conformal score $c(x_i, y_i)$ is thus composed of this cumulative probability and the predicted probability of class $y_i$.

\paragraph{Proposal} Our approach differs from prior work in that the conformal score is derived by transforming the probabilities output by $f$ using test-time augmentation. Concretely, this replaces Equation~\ref{eq:1} with the following, parametrized by augmentation policy $\mathcal{A}$ and augmentation weights $\theta$. 

\begin{align}
\hat{\pi}_x(y') &= \hat{p}(y = y' | x_i) = g(x_i; f, \mathcal{A}, \theta)
\end{align}

Aggregation weights $\theta$ are applied to the logits output by classifier $f$, and transformed to be proper probabilities by applying a softmax function. We learn the aggregation weights $\theta$ using a portion of the validation set, $D^{(TTA)}$, distinct from calibration set used to identify the conformal threshold ($D^{(cal)}$). In contrast to traditional approaches, where all labeled data is used to estimate the conformal threshold, we instead reserve a portion to learn the test-time augmentation policy. 

We learn a set of weights which maximize classification accuracy on $D^{(TTA)}$ by minimizing the cross-entropy loss\footnote{We found no significant improvement by using alternate losses considered in the conformal prediction literature (e.g. focal or conformal training loss). See Table \ref{tab:alternate-losses} in the Appendix.} computed between the predicted probabilities and true labels. More formally, $g$ applies $\theta$ and $\mathcal{A}$ as follows: 

\begin{eqnarray}
    g(x_i ; f, \mathcal{A}, \Theta) &=& \sigma ( \theta^T \mathbf{A}(f, \mathcal{A}, x_i) )
\end{eqnarray}

where $\mathbf{A}$ uses $f$ to map input $x_i$ to a $M \times K$ matrix of predicted logits where $M$ is the number of augmentations and $K$ is the number of classes. $\theta$ is a $1 \times m$ vector corresponding to augmentation-specific weights. Each row in $\mathbf{A}(f, \mathcal{A}, x_i)$ represents the pre-trained classifier's predicted logits on augmentation $a_m$ of $x_i$. TTA-Learned refers to TTA combined with learned augmentation weights, while TTA-Avg refers to a simple average over the augmentations.

We refer to the fraction of the validation set allotted to $D^{(TTA)}$ as $\beta$. Figure \ref{fig:supp_tta_data} shows that performance is not sensitive to the choice of $\beta$; as a result, all experiments use $\beta = .2$ (see supplement for further discussion). This does reduce the amount of data available to identify the appropriate threshold, but we find that the benefits TTA confers outweigh the cost to threshold estimation. Computational cost scales linearly with the size of $\mathcal{A}$; each additional augmentation translates to a forward pass of the base classifier. One can use the learned weights to save computation by identifying which test-time augmentations to generate.

\paragraph{Preserving exchangeability} The validity of conformal prediction depends upon the assumption of exchangeability: that all orderings of examples are equally likely (in effect, meaning that the distribution of examples in the calibration set is indistinguishable from the distribution of unseen examples). Prior work has shown that the conformal procedure is valid under deterministic transformations \citep{kuchibhotla2020exchangeability}; by using distinct examples to learn the test-time augmentation policy, the proposed approach constitutes a deterministic transformation applied to both the calibration set and unseen examples. If we were to instead use the \emph{same} examples to learn the test-time augmentation policy and the conformal threshold, exchangeability could be broken. 



\section{Experimental Set-Up}
\label{sec:set-up}

\paragraph{Datasets} We show results on the test splits of three widely used image classification datasets: ImageNet \citep{deng_imagenet_2009} (50,000 natural images spanning 1,000 classes), iNaturalist \citep{van_horn_benchmarking_2021} (100,000 images spanning 10,000 species), and CUB-Birds \citep{wah2011caltech} (5,794 images spanning 200 categories of birds). Images are distributed evenly over classes in ImageNet and iNaturalist, while CUB-Birds has between 11 and 30 images per class.

\paragraph{Models} The default model architecture, across all datasets, is ResNet-50 \citep{he_deep_2016}. The accuracies of the base classifiers are  76.1\% (ImageNet), 76.4\% (iNaturalist), and 80.5\% (CUB-Birds). To study the relationship between model complexity and performance, we also provide results using ResNet-101 and ResNet-152 on ImageNet. For ImageNet, we make use of the pretrained models made available by PyTorch \citep{paszke_pytorch_2019}. For iNaturalist, we use a model made public by \citet{niers_tom_inaturalist_competition_2021}. For CUB-Birds, we train a network by finetuning the final layer of a ResNet-50 model initialized with ImageNet's pretrained weights.

\paragraph{Augmentations} We consider two augmentation policies. The first (the \emph{simple} augmentation policy) consists of a random-crop  and a horizontal-flip; to produce a random crop, we pad the original image with 4 pixels and take a 256x256 crop of the expanded image (thereby preserving the original image resolution). The simple augmentation policy is widely used because its augmentations are likely to be label-preserving. The second, which we refer to as the \emph{expanded} augmentation policy, consists of 12 augmentations: increase-sharpness, decrease-sharpness, autocontrast, invert, blur, posterize, shear, translate, color-jitter, random\_crop, horizontal-flip, and random-rotation. The supplement contains a description of each augmentation. These augmentations are not always label preserving, but, as we show, can improve performance when weights are learned.

\paragraph{Baselines} We benchmark results using two conformal scores (translating to different definitions of $c(x,y)$ in Equation \ref{eqn:c_score}). The first score is APS \citep{romano_classification_2020} (described in Eqn. \ref{eqn:c_score}), which represents the cumulative probability required to include the correct class, and the second is RAPS \citep{angelopoulos_uncertainty_2022}, which modifies APS by adding a term to penalize large set sizes. 
For all experiments, we do not allow sets of size 0. 
We implement RAPS and APS using code provided by \citet{angelopoulos_uncertainty_2022}, and automatically select hyperparameters $k_{reg}$ and $\lambda$ to minimize set size. 
We also compare against conformal prediction using a simple average over the test-time augmentations (TTA-Avg). In the supplement, we also compare against non-conformal Top-1 and Top-5 prediction sets.

\paragraph{Evaluation} We evaluate results using the three metrics commonly used in the conformal prediction literature: efficiency, coverage, and adaptivity. We quantify efficiency using both average prediction set size (measured across all examples) and class-conditional prediction set size (measured across all examples in a class). Coverage is the percentage of sets containing the true label. We define adaptivity as the size-stratified coverage violation (SSCV), introduced by \citet{angelopoulos_uncertainty_2022}. We first partition examples based upon the size of the prediction set. We create bins for set sizes of $[0, 1], [2, 3], [4, 10], [11, 100], \text{and } [101,]$. We then compute the empirical coverage within each bin, and compute adaptivity as the maximum difference between theoretical coverage and empirical coverage across bins. The closer this value is to 0, the better the adaptivity.

For each dataset, we report results across 10 randomly generated splits into validation and test sets. For all experiments (save for the validation set size experiment), the validation set and test set are the same size. We allot 20\% of examples from the validation set to $D^{(TTA)}$ (used to learn TTA policy), and allot the remaining examples to the calibration set. For the experiment studying validation set size, we downsample the validation set. We compute statistical significance using a paired t-test, with a Bonferroni correction \citep{weisstein2004bonferroni} for multiple hypothesis testing. Code to reproduce all experiments will be made publicly available.

\renewcommand{\arraystretch}{1.3}

\begin{table*}[t]
\centering
\begin{minipage}[t]{\textwidth}
\resizebox{1\textwidth}{!}{%
\rowcolors{2}{gray!20}{white}
\begin{tabular}{cl|c|c|c|c|c|c}
\toprule
  &  & \multicolumn{3}{c|}{Expanded Augmentation Policy} & \multicolumn{3}{c}{Simple Augmentation Policy} \\
\toprule
 Alpha & Method & ImageNet & iNaturalist & CUB-Birds & ImageNet & iNaturalist & CUB-Birds \\
\toprule
\hline
0.01 & RAPS & 37.751 $\pm$ 2.334 & 61.437 $\pm$ 6.067 & \textbf{15.293 $\pm$ 2.071} & 37.751 $\pm$ 2.334 & 61.437 $\pm$ 6.067 & \textbf{15.293 $\pm$ 2.071}\\

0.01 & RAPS+TTA-Avg & 35.600 $\pm$ 2.200 & 57.073 $\pm$ 5.914 & \textbf{13.111 $\pm$ 2.470} & \textbf{31.681 $\pm$ 3.057} & \textbf{54.169 $\pm$ 6.319} & \textbf{14.550 $\pm$ 1.425}\\

0.01 & RAPS+TTA-Learned & \textbf{31.248 $\pm$ 2.177} & \textbf{53.195 $\pm$ 4.884} & \textbf{14.045 $\pm$ 1.323} & \textbf{32.702 $\pm$ 2.409} & \textbf{51.391 $\pm$ 5.211} & \textbf{13.803 $\pm$ 1.734}\\

\hline
0.05 & RAPS & 5.637 $\pm$ 0.357 & \textbf{7.991 $\pm$ 1.521} & 3.624 $\pm$ 0.361 & 5.637 $\pm$ 0.357 & 7.991 $\pm$ 1.521 & 3.624 $\pm$ 0.361\\

0.05 & RAPS+TTA-Avg & 5.318 $\pm$ 0.113 & \textbf{7.067 $\pm$ 0.344} & \textbf{3.116 $\pm$ 0.210} & \textbf{4.908 $\pm$ 0.099} & \textbf{6.451 $\pm$ 0.279} & \textbf{3.249 $\pm$ 0.307}\\

0.05 & RAPS+TTA-Learned & \textbf{4.889 $\pm$ 0.168} & \textbf{6.682 $\pm$ 0.447} & \textbf{3.571 $\pm$ 0.576} & \textbf{5.040 $\pm$ 0.176} & \textbf{6.788 $\pm$ 0.496} & \textbf{3.290 $\pm$ 0.186}\\

\hline
0.10 & RAPS & 2.548 $\pm$ 0.074 & 2.914 $\pm$ 0.116 & 2.038 $\pm$ 0.153 & 2.548 $\pm$ 0.074 & 2.914 $\pm$ 0.116 & 2.038 $\pm$ 0.153\\

0.10 & RAPS+TTA-Avg & 2.470 $\pm$ 0.071 & 2.740 $\pm$ 0.026 & \textbf{1.780 $\pm$ 0.139} & \textbf{2.327 $\pm$ 0.086} & \textbf{2.610 $\pm$ 0.031} & \textbf{1.881 $\pm$ 0.118}\\

0.10 & RAPS+TTA-Learned & \textbf{2.312 $\pm$ 0.054} & \textbf{2.625 $\pm$ 0.043} & \textbf{1.893 $\pm$ 0.187} & \textbf{2.362 $\pm$ 0.065} & 2.638 $\pm$ 0.026 & \textbf{1.840 $\pm$ 0.106}\\

\hline
\end{tabular}

}
\caption{\textbf{Reductions in prediction set size across datasets, augmentation policies, and coverage specifications.} Each entry corresponds to the average prediction set size across 10 calibration/test splits. Bolded entries represent performance that is either (a) significantly better compared to the baseline (RAPS), or (b) indistinguishable from the best approach. Table \ref{results-raps-coverage} reports achieved coverage. Corresponding results for APS can be found in Table \ref{results-aps-coverage}.}
\label{results-raps}
\end{minipage}
\end{table*}

\begin{figure*}
\begin{center}
\includegraphics[width=\textwidth]{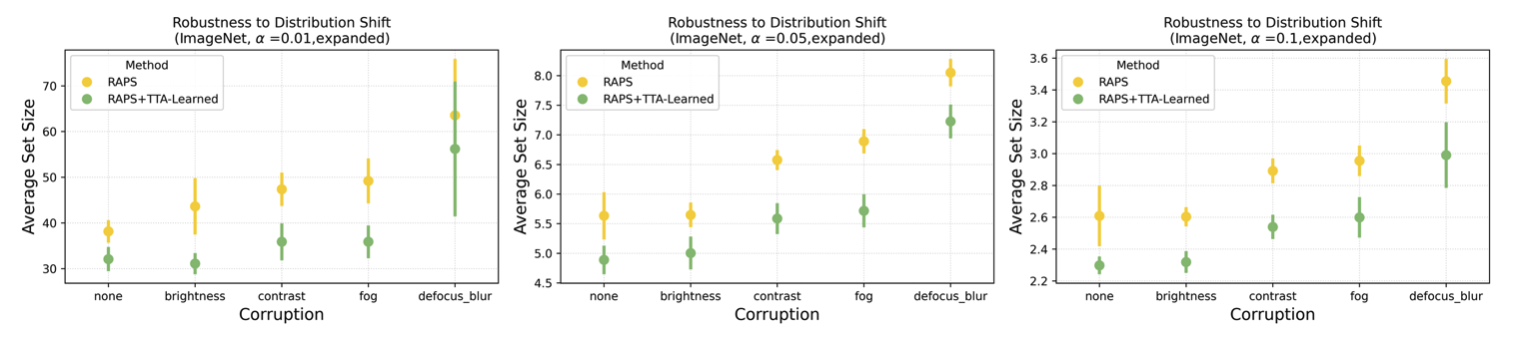}
\end{center}
\caption{\textbf{Robustness to distribution shift.}  We compare average prediction set size achieved by RAPS (yellow) to average prediction set size achieved when combining RAPS with TTA-Learned (green). Results reflect the distribution of average prediction set size across 10 runs using ImageNet and ResNet50. We evaluate performance on different corruptions (x-axis) and different coverage guarantees (left, middle, right). RAPS+TTA-Learned (green) produces a noticeable reduction in prediction set size, even when subject to distribution shift, with no loss in coverage. Refer to Figure \ref{fig:supp_coverage_robustness} in the supplement for a comparison of coverage achieved by both methods.}
\label{fig_robustness}
\end{figure*}

\renewcommand{\arraystretch}{1.35}

\begin{table*}
\centering

\resizebox{\textwidth}{!}{%
\rowcolors{2}{gray!20}{white}
\begin{tabular}{cl|c|c|c|c|c|c}
\toprule
  &  & \multicolumn{3}{c|}{Expanded Augmentation Policy} & \multicolumn{3}{c}{Simple Augmentation Policy} \\
\toprule
 Alpha & Method & ResNet-50 & ResNet-101 & ResNet-152 &  ResNet-50  & ResNet-101 & ResNet-152 \\
\toprule
\hline
0.01 & RAPS & 37.751 $\pm$ 2.334 & 33.624 $\pm$ 1.796 & 29.560 $\pm$ 3.481 & 37.751 $\pm$ 2.334 & 33.624 $\pm$ 1.796 & 29.560 $\pm$ 3.481\\

0.01 & RAPS+TTA-Avg & 35.600 $\pm$ 2.200 & 30.220 $\pm$ 1.774 & 27.203 $\pm$ 2.526 & \textbf{31.681 $\pm$ 3.057} & \textbf{27.206 $\pm$ 1.840} & \textbf{24.106 $\pm$ 2.100}\\

0.01 & RAPS+TTA-Learned & \textbf{31.248 $\pm$ 2.177} & \textbf{25.722 $\pm$ 1.713} & \textbf{23.615 $\pm$ 1.656} & \textbf{32.702 $\pm$ 2.409} & \textbf{26.760 $\pm$ 1.974} & \textbf{24.765 $\pm$ 2.736}\\

\hline
0.05 & RAPS & 5.637 $\pm$ 0.357 & 4.785 $\pm$ 0.102 & 4.376 $\pm$ 0.078 & 5.637 $\pm$ 0.357 & 4.785 $\pm$ 0.102 & 4.376 $\pm$ 0.078\\

0.05 & RAPS+TTA-Avg & 5.318 $\pm$ 0.113 & 4.433 $\pm$ 0.137 & 4.163 $\pm$ 0.185 & \textbf{4.908 $\pm$ 0.099} & \textbf{4.147 $\pm$ 0.122} & \textbf{3.868 $\pm$ 0.126}\\

0.05 & RAPS+TTA-Learned & \textbf{4.889 $\pm$ 0.168} & \textbf{4.200 $\pm$ 0.200} & \textbf{3.824 $\pm$ 0.128} & \textbf{5.040 $\pm$ 0.176} & \textbf{4.194 $\pm$ 0.194} & \textbf{3.916 $\pm$ 0.356}\\

\hline
0.10 & RAPS & 2.548 $\pm$ 0.074 & 2.267 $\pm$ 0.024 & 2.109 $\pm$ 0.027 & 2.548 $\pm$ 0.074 & 2.267 $\pm$ 0.024 & 2.109 $\pm$ 0.027\\

0.10 & RAPS+TTA-Avg & 2.470 $\pm$ 0.071 & 2.164 $\pm$ 0.031 & 2.049 $\pm$ 0.028 & \textbf{2.327 $\pm$ 0.086} & \textbf{2.093 $\pm$ 0.035} & \textbf{1.996 $\pm$ 0.018}\\

0.10 & RAPS+TTA-Learned & \textbf{2.312 $\pm$ 0.054} & \textbf{2.099 $\pm$ 0.040} & \textbf{1.993 $\pm$ 0.026} & \textbf{2.362 $\pm$ 0.065} & \textbf{2.091 $\pm$ 0.041} & \textbf{1.988 $\pm$ 0.020}\\

\hline
\end{tabular}
}
\caption{\textbf{Reductions in prediction set size across base classifiers on ImageNet.} TTA-Learned can bridge the performance gap between different classifiers (for example, outperforming ResNet-152 alone when combined with ResNet-101), and yields significant reductions in set size regardless of the pretrained classifier used. We report achieved coverage in Table \ref{model-results-raps-coverage}.}
\label{model-results-raps}
\end{table*}

\section{Results}

We compare against RAPS, which outperformed other baselines in every experiment . (We provide results comparing our method to APS and the Top-K baselines in the supplement. Variants of each experiment across multiple $\alpha$ and datasets are also in the supplement.)
We then examine the dependence of these results on dataset, base model, and class. We conclude by providing intuition about why test-time augmentation improves the efficiency of conformal predictors.

\subsection{Reductions in prediction set size}
\label{sec:tta-general-results}

We begin with results in the context of the expanded augmentation policy.
Learned test-time augmentation policies produce meaningfully significant reductions in prediction set size (RAPS+TTA-Learned in Table \ref{results-raps} and APS+TTA-Learned in Table \ref{results-aps}). TTA-Learned reduces prediction set sizes significantly in 16 of the 18 cases, and performs comparably in the remaining 2.
Across all cases, the combination of RAPS, TTA-Learned, and the expanded augmentation policy produces the smallest average set sizes.

TTA-Learned performs comparably or better than TTA-Avg in all comparisons.
Certain augmentations in the expanded augmentation policy (blur, decrease sharpness, and shear) are consistently assigned a weight of 0, while certain augmentations are consistently included in learned policies (autocontrast, translate). Augmentations assigned zero weight provide no additional information about the true label (for example, they may not preserve the label within the image, or they may be redundant with other augmentations included in the policy).

While TTA improves both RAPS and APS, the improvements are larger for APS. This is because TTA, like RAPS, tempers the predicted probabilities. TTA lowers the maximum predicted probability on average, thereby reducing model overconfidence. Consequently, the predicted probability assigned to the remaining classes is higher. This is why the expanded augmentation policy demonstrates stronger performance than the simple augmentation policy: it tempers the probabilities to a greater extent.

TTA-Learned preserves coverage across all experiments, since it respects the assumption of exchangeability. In some cases, TTA significantly improves coverage, although the magnitude of this difference is small (results can be found in Tables \ref{results-raps-coverage} and \ref{model-results-raps-coverage}).

We next evaluate adaptivity using size stratified coverage violation (SSCV). At low alpha ($\alpha$ = .01, and $\alpha$ = .05), TTA-Learned improves efficiency without diminishing adaptivity (Table \ref{adaptivity-results}).  

\subsection{Robustness to distribution shift}

Next, we evaluate the performance of test-time augmented conformal prediction on out-of-distribution examples. 
 While conformal prediction does not guarantee coverage in these settings, distribution shifts are ubiquitous in practice \citep{koh2021wilds}. Empirical performance is thus of practical interest. 
The training procedures for both test-time augmented conformal prediction and conformal prediction remain the same and use in-distribution examples from ImageNet and a ResNet50 classifier. We evaluate each conformal predictor on four types of image corruptions drawn from ImageNet-C \citep{hendrycks2019benchmarking}. Figure \ref{fig_robustness} plots the results; across all corruptions and coverage guarantees, test-time augmented conformal prediction produces smaller prediction sets than conformal prediction alone. Importantly, test-time augmented conformal prediction achieves this with \emph{no} loss of coverage (Figure \ref{fig:supp_coverage_robustness}).

\subsection{Datasets, augmentation policies, and models}
\label{sec:dependencies}

\paragraph{Dependence on dataset} TTA consistently improves prediction set sizes on ImageNet and iNaturalist, but not on CUB-Birds. This may be because the validation set size for CUB-Birds (2,827 images) is an order of magnitude smaller than the validation sets for ImageNet (25,000 images) and iNaturalist (50,000 images). This is consistent with our finding that effectiveness of TTA is positively correlated with the size of the validation set (Figure \ref{fig:cal_set_size}).

\paragraph{Dependence on augmentation policy} We find that the expanded augmentation policy produces larger reductions in set size than the simple augmentation policy, primarily at low $\alpha$. This is in spite of the fact that the expanded augmentation policy contains many augmentations outside of the base classifier's training-time augmentation policy. 
When we vary the number of augmentations included in an augmentation policy, we see that larger augmentation policies also yield greater reductions in average prediction set size (Figure \ref{fig:aug_policy}). 
That said, the simple augmentation policy does have its place; it requires fewer forward passes during inference. In the absence of a learned aggregation function, our results suggest that aggregating using an average can still improve the efficiency of conformal predictors (outperforming the original conformal score in 11 comparisons, matching performance in 3, and under-performing in 4).

\begin{figure*}
\begin{center}
\includegraphics[width=\textwidth]{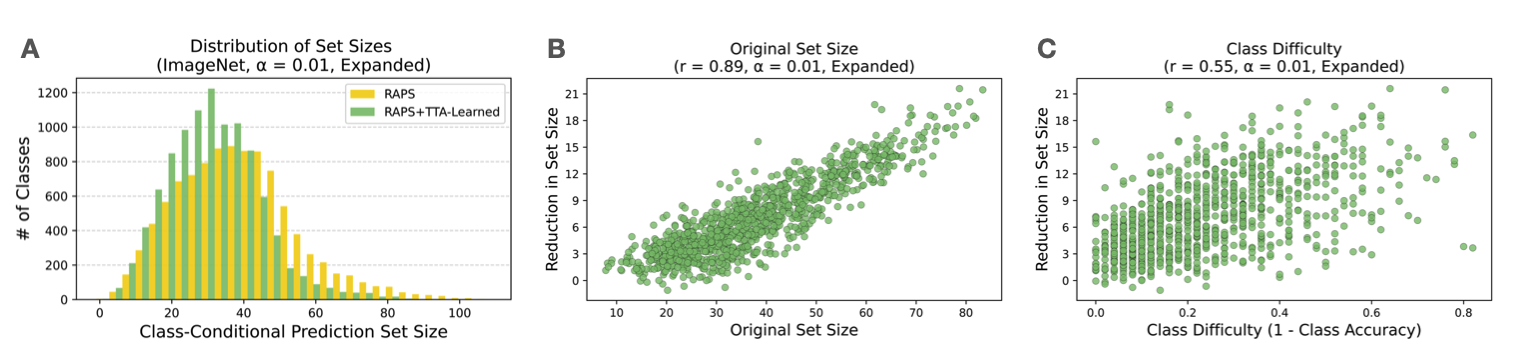}
\end{center}
\caption{(A) \textbf{Class-conditional prediction set sizes.}  We plot the distribution of class-conditional prediction set sizes, for ImageNet and ResNet-50 with $\alpha = .01$. RAPS+TTA-Learned (green) produces a noticeable reduction in class-conditional prediction set sizes. (B, C) \textbf{Relationship between TTA improvements and original class set sizes and  class difficulty}. TTA introduces the largest improvements for classes with the largest original prediction set sizes (B) and classes on which the underlying classifier is often incorrect (C). Each point represents the average prediction set size for each class, across 10 splits. 
}
\label{fig_class_specific}
\end{figure*}

\begin{figure*}[t!]
\begin{center}
\includegraphics[width=1\textwidth]{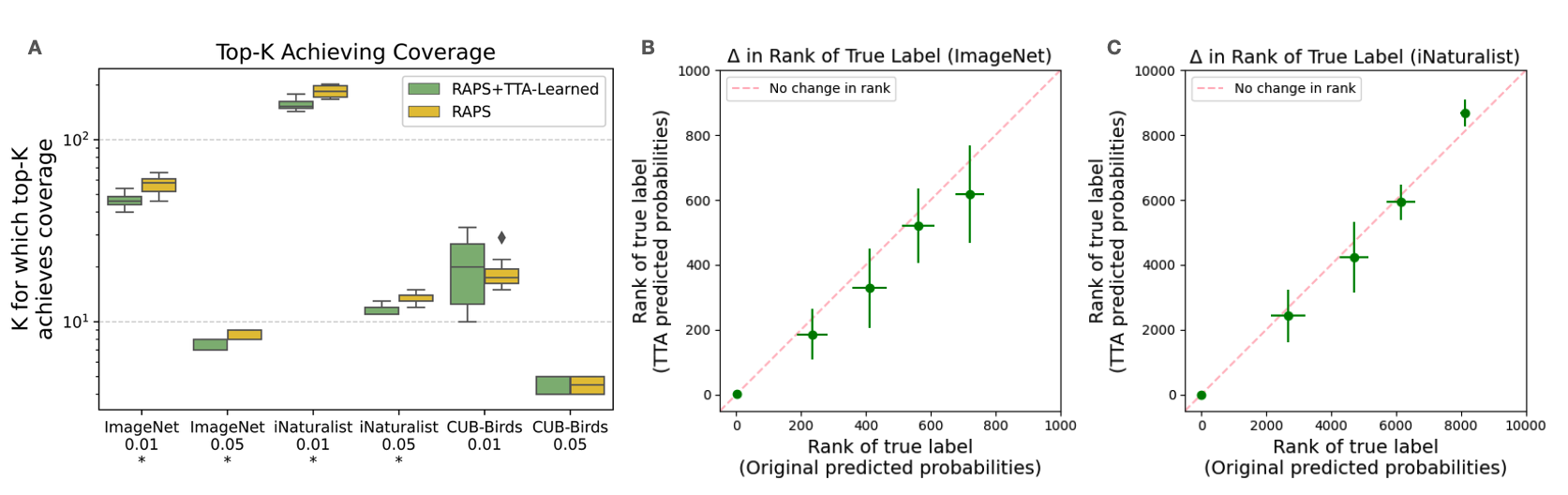}
\end{center}
\caption{
(A) \textbf{Effect of TTA-Learned on optimal Top-K}: TTA-Learned significantly lowers the value of k required for Top-k prediction sets to achieve coverage on ImageNet and iNaturalist, but not on CUB-Birds. (B,C) \textbf{Effect of TTA-Learned on rank of true class}: TTA-Learned improves the rank of the true class among the sorted predicted probabilities for a given example for both ImageNet (B) and iNaturalist (C). We plot the rank using the original predicted probabilities compared to the TTA-transformed probabilities, binning all examples in the validation set into five equal-width bins. 
Dots that fall below the red line indicate that TTA improves the rank of the true class.  
}
\label{fig_aug_cal_kreg}
\end{figure*}

\paragraph{Dependence on base model} We tested the generalizability of our results to other models by rerunning the ImageNet experiments using ResNet-101 (accuracy of 77.4\%) and ResNet-152 (accuracy of 78.3\%). Unsurprisingly, more accurate models result in smaller prediction set sizes (Table \ref{model-results-raps}). TTA variants of conformal prediction again produce significant improvements in set size while maintaining coverage. We note that the combination of TTA with  ResNet-101 produces smaller set sizes than the more complex ResNet-152 alone. For example, when $\alpha$ is set to .01, RAPS+TTA-Learned and ResNet-101 produce set sizes that contain, on average, 26.5 classes, while RAPS and ResNet-152 produce an average set size of 29.6.

\subsection{Class-Specific Analysis}
\label{sec:class-specific}

We have established that on average TTA is a useful addition to the conformal pipeline. We now investigate the source of this improvement.  We make two empirical observations. First, classes with larger predicted set sizes benefit most from the introduction of TTA. Figure \ref{fig_class_specific} shows that a class's average prediction set size is significantly correlated with the change in set size produced by TTA-Learned (with the expanded augmentation policy and $\alpha = .01$, $r = 0.89$, and $p < 1e-10$). Second, we find that class difficulty is significantly associated with changes in set size introduced by TTA (with the expanded augmentation policy and $\alpha = .01$, r = 0.55 and \emph{p} $<$ 1e-10). Prediction sets for classes that are difficult to predict benefit more from TTA compared to their easier counterparts. These observations are related; harder classes receive larger set sizes, and consequently, offer larger room for improvements in efficiency.

\subsection{Intuition} Why does the addition of test-time augmentation produce smaller prediction set sizes? In short, TTA improves top-K accuracy. We verify this claim by estimating $k$ such that the uncertainty sets comprised of the top $k$ predicted classes for each example achieve a marginal coverage of $(1-\alpha)$. We see that for those datasets (ImageNet and iNaturalist) where TTA produces significant reductions in set size, TTA-transformed predictions--both with a simple average and learned weights-- produce significantly lower values for $k$ compared to the original predictions (Figure \ref{fig_aug_cal_kreg}A). This is \emph{not} true for CUB-Birds, on which TTA offers little to no improvement. One could use such a procedure to determine whether TTA is worth adding to a conformal pipeline without collecting labeled data beyond the calibration dataset.

Another way to understand the impact of TTA is to consider the effect on the \emph{ordering} of classes. It has been observed in the test-time augmentation literature that TTA often promotes the true class from the second-highest to the highest predicted probability, thereby correcting the classification. This finding does not fully explain the value of test-time augmentation to conformal prediction since only 3\% of the prediction sets which reduce in size are associated with a corrected Top-1 classification. What TTA did do was increase the predicted probability of the true class \emph{even when it is predicted to be unlikely} (for example, promoting the true class from 200th most likely to 100th most likely).  We visualize this effect in Figure \ref{fig_aug_cal_kreg} by plotting the change in true class rank (the index at which the true class appears in the sorted list of predicted probabilities) for all examples in the  validation set, stratified into 5 equal-width bins. The lower left point captures examples that are classified correctly; here, test-time augmentation introduces little to no change. In subsequent bins, we see that TTA typically promotes the rank of the true class. We also include the standard deviation across the true class ranks in the original predicted probabilities (x-axis) and the TTA-transformed probabilities (y-axis).

\section{Limitations}
Learned test-time augmentation policies require two ingredients: labeled data and multiple forward passes. 
Although one can minimize costs by parallelizing computation or by using the learned weights to identify which augmentations to generate, inference will always cost more with test-time augmentation. 
Our results are also limited to image classification. We do not consider other modalities, for which appropriate transformations will substantially differ. Future work should consider how these results may generalize to non-vision tasks. Finally, test-time augmentation is one approach to generating ensembles in conformal prediction. Many other more computationally expensive approaches exist. The tradeoff between computation and ensemble performance remains a useful avenue for future work.

\section{Conclusion}

We show that test-time augmented conformal prediction produces smaller sets than conformal prediction alone.
The proposed approach is effective, efficient, and simple: it reduces prediction set sizes by up to 30\%, requires no model re-training, and relies on a portion of labeled data already available to split conformal predictors.
Our experiments also indicate that test-time augmented conformal prediction exhibits greater efficiency under four common corruption-based distribution shifts. 
Test-time augmentation is able to do this by improving the underlying classifier's robustness to domain-specific invariances, in the form of data augmentation. 
Efforts to improve the efficiency of conformal predictors could, as a first step, aim to improve the robustness of the underlying classifier.
 The performance of TTA-Learned also suggests that there are settings in which it is wise to use a portion of the labeled data to improve the underlying model, instead of reserving all labeled data for the calibration set.
In sum, our work takes a step towards practically useful conformal predictors by improving efficiency, without sacrificing adaptivity or coverage.

{
    \small
    \bibliographystyle{ieeenat_fullname}
    \bibliography{main}
}

\clearpage
\newpage

\setcounter{section}{0}
\renewcommand{\thesection}{S\arabic{section}}

\setcounter{figure}{0}
\renewcommand{\figurename}{Figure}
\renewcommand{\thefigure}{S\arabic{figure}}
\setcounter{table}{0}
\renewcommand{\tablename}{Table}
\renewcommand{\thetable}{S\arabic{table}}

\section*{Supplementary Material}

\section{Experimental Details}
\subsection{Augmentations}
\label{supp:augs}

The simple augmentation policy consists of a random crop and a horizontal flip, drawn from a widely used test-time augmentation policy in image classification \cite{krizhevsky_imagenet_2012}. The random crop pads the original image by 4 pixels and takes a 256x256 crop of the resulting image. The expanded augmentation consists of 12 augmentations; certain augmentations are stochastic, while others are deterministic. We design this set based on the augmentations included in AutoAugment \citep{cubuk_autoaugment_2019}. We exclude certain augmentations, however, to exclude 1) redundancies among augmentations and thereby make the learned weights interpretable and 2) augmentations are unlikely to be label-preserving. In particular, we exclude CutOut (because it is clearly not label-preserving in many domains) and exclude brightness, contrast, saturation, and color for their overlap with color-jitter. We also exclude contrast, because it is already modified via autocontrast, and equalize and solarize for their overlap with autocontrast and invert. This leaves us the following augmentations:
\\
\begin{itemize}
\small
    \itemsep 0em
    \item \emph{Shear}: Shear an image by some number of degrees, sampled between [-10, 10] (stochastic).
    \item \emph{Translate}: Samples a vertical shift (by fraction of image height) from [0, .1] (stochastic).
    \item \emph{Rotate}: Samples a rotation (by degrees) from [-10, 10] (stochastic).
    \item \emph{Autocontrast}: Maximizes contrast of images by remapping pixel values such that the the lowest becomes black and the highest becomes white (deterministic).
    \item \emph{Invert}: Inverts the colors of an image (deterministic).
    \item \emph{Blur}: Applies Gaussian blur with kernel size 5 (and default $\sigma$ range of [.1, .2]) (stochastic). 
    \item \emph{Posterize}: Reduces the number of bits per channel to 4 (deterministic).
    \item \emph{Color Jitter}: Randomly samples a brightness, contrast, and saturation adjustment parameter from the range [.9, 1.1] (stochastic). 
    \item \emph{Increase Sharpness}: Adjusts sharpness of image by a factor of 1.3 (deterministic).
    \item \emph{Decrease Sharpness}: Adjusts sharpness of image by a factor of 0.7 (deterministic).
    \item \emph{Random Crop}: Pads each image by 4 pixels, takes a 256x256 crop, and then proceeds to take a 224x224 center crop (stochastic).
    \item \emph{Horizontal Flip}: Flips image horizontally (deterministic).
\end{itemize}

There are many possible expanded test-time augmentation policies; this particular policy serves as an illustrative example.

\subsection{Learning aggregation function}
\label{supp:agg_training}

We learn $\hat{g}$ by minimizing the cross-entropy loss with respect to the true labels on the calibration set. Specifically, we learning the weights using SGD with a learning rate of .01, momentum of .9, and weight decay of 1e-4. We train each model for 50 epochs. There are natural improvements to our optimization, but this is not the focus of our work. Instead, our goal is to highlight the surprising effectiveness of TTA-Learned \emph{without} the introduction of hyperparameter optimization. We train all models using a machine equipped with 4 Titan Xp GPUs, 2 Octa Intel Xeon E5-2620 CPUs, and 1TB of RAM.

\section{Supplementary Results}

\subsection{Test-Time Augmentation and APS}
\label{supp:aps}

TTA-Learned combined with the expanded augmentation policy produces the smallest set sizes when combined with APS, across the datasets considered (Table \ref{results-aps})  and each base classifier (Table \ref{model-results-aps}). In contrast to the results using RAPS, TTA-Learned does not significantly outperform TTA-Avg when combined with APS. The central reason is that the improvements TTA confers --- namely, improved top-k accuracy --- do not address the underlying sensitivity of APS to classes with low predicted probabilities. As \citet{angelopoulos_uncertainty_2022} discuss, APS produces large prediction sets because of noisy estimates of small probabilities, which then end up included in the prediction sets. Both TTA-Learned and TTA-Avg smooth the probabilities: they reduce the number of low-probability classes by aggregating predictions over perturbations of the image. The benefit that both TTA-Learned and TTA-Avg add to APS is thus similar to how RAPS penalizes classes with low probabilities.

\subsection{Comparison to Top-1 and Top-5}
\label{supp:topk}

We expand Table \ref{results-raps} to include the Top-1 and Top-5 baselines in Table \ref{results-topk}. Unsurprisingly, neither outperform RAPS, and consequently none outperform the combination of RAPS, TTA-Learned, and the expanded augmentation policy.

\subsection{Comparison to minimizing focal loss}
\label{supp:focal_loss}

We expand Table \ref{results-raps} to include results for a variant of TTA-Learned which uses a focal loss in place of the cross-entropy loss. We conduct this exploration because empirically, the focal  loss has  been known to produce better-calibrated models. Table \ref{tab:alternate-losses} reports our results. We see little difference between results when using a different loss function; RAPS+TTA-Leanred still outperforms RAPS + an average over the test-time augmentations, and RAPS alone. While this speaks to the method's flexibility to different loss functions, it is possible that the use of a loss function designed to reduce prediction set size could produce better performance.

\subsection{Impact on coverage}

We provide exact values of coverage for the main experiments here. In short, TTA-Learned combined with the expanded augmentation policy \emph{never} worsens coverage, and in some cases, significantly improves it (although the improvements are small in magnitude). Coverage values for the RAPS experiment across coverage values and datasets can be found in Table \ref{results-raps-coverage} and coverage values for the RAPS experiment across base classifiers  can be found in Table \ref{model-results-raps-coverage}. Similarly, we provide coverage values for the APS experiment across datasets (Table \ref{results-aps-coverage}) and across models (Table \ref{model-results-aps-coverage}).

\subsection{Impact of different coverage guarantees and datasets}
\label{supp:replicates}
\label{supp:class_specific}

\begin{figure*}
\begin{center}
\includegraphics[width=\textwidth]{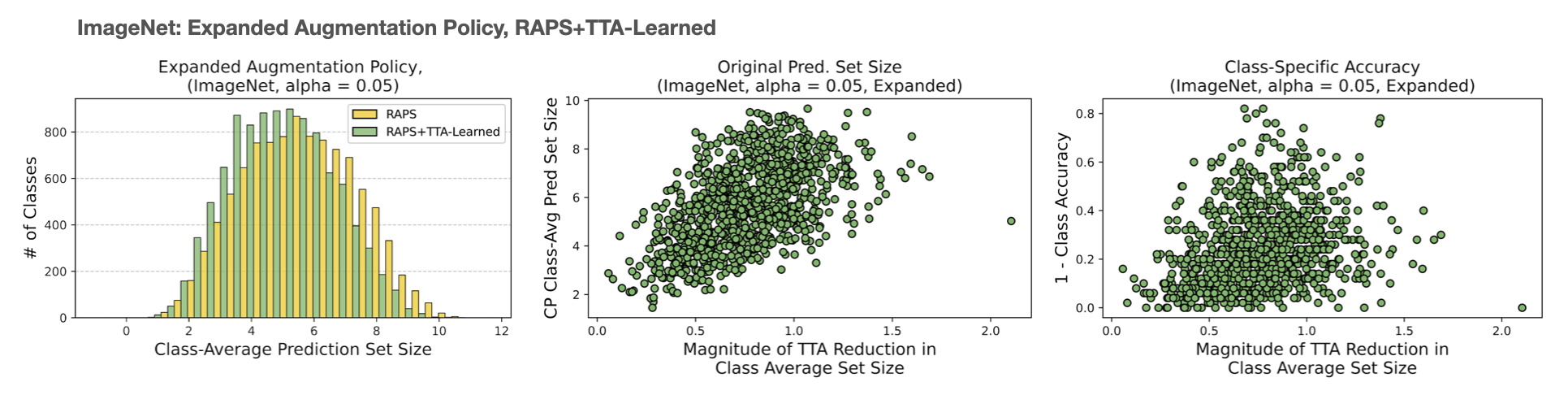}
\end{center}
\vspace{-1em}
\caption{\textbf{Class-specific performance for ImageNet}, for a coverage of 95\% $\alpha = .05$. Using the expanded augmentation policy RAPS+TTA-Learned produces a noticeable shift in class-average prediction set sizes to the left. There is a significant correlation between original prediction set size and improvements from TTA (middle) and between class difficulty and improvements from TTA (right).}
\label{fig:class_specific_imagenet_.05_expanded}
\vspace{-.5em}
\end{figure*}

\begin{figure*}
\begin{center}
\includegraphics[width=\textwidth]{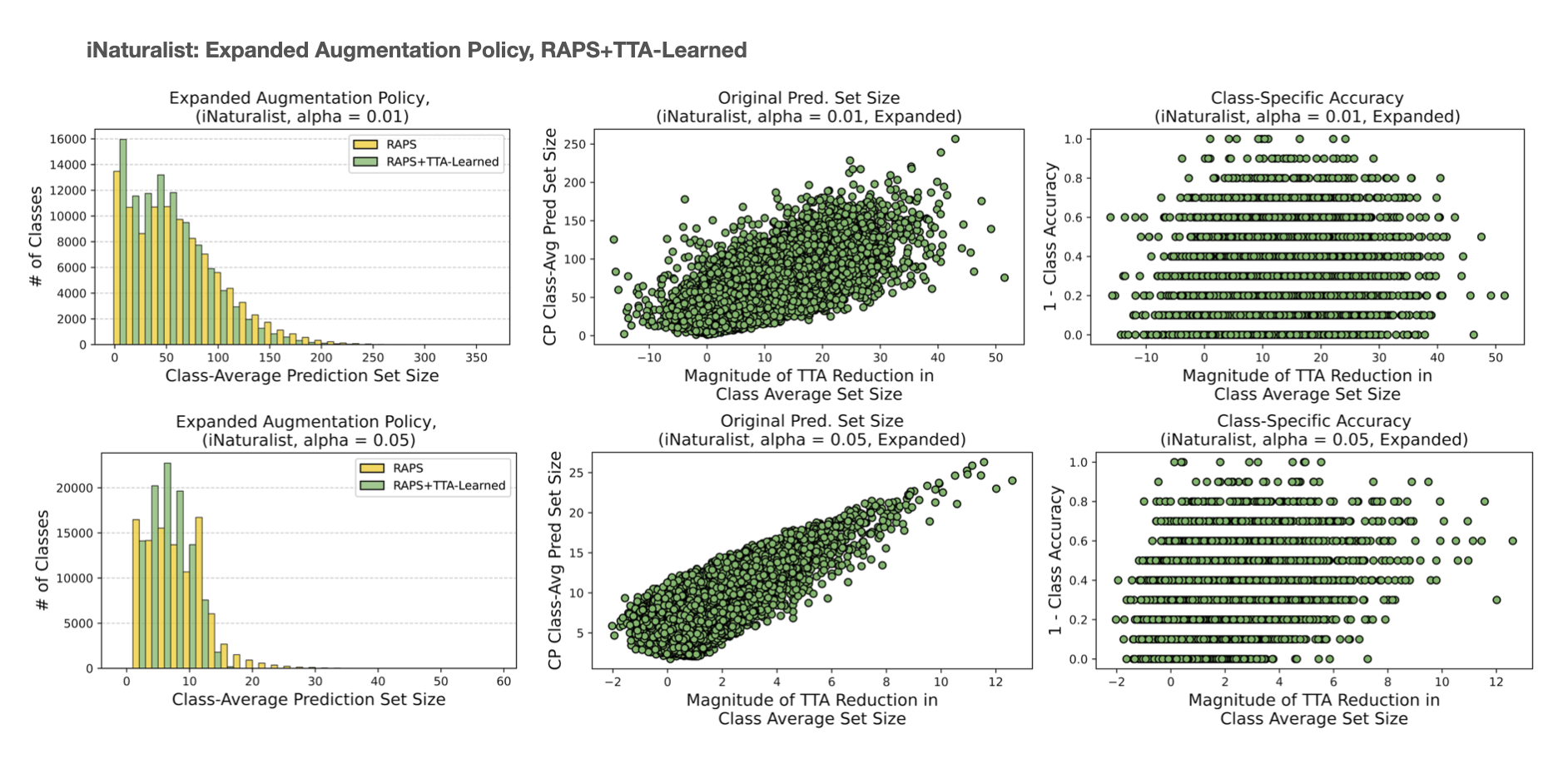}
\end{center}
\vspace{-1em}
\caption{\textbf{Class-specific performance for iNaturalist}, for $\alpha = .01$ (top) and $\alpha = .05$ (bottom).  We see a consistent relationship between TTA improvements and original class-average prediction set size (middle) and class difficulty (right). Estimates of class-specific accuracy on iNaturalist are quite noisy because there are 10 images per class (which produces distinct accuracy bands).}
\label{fig:class_specific_inat_expanded}\vspace{-.5em}
\end{figure*}

\begin{figure*}
\begin{center}
\includegraphics[width=\textwidth]{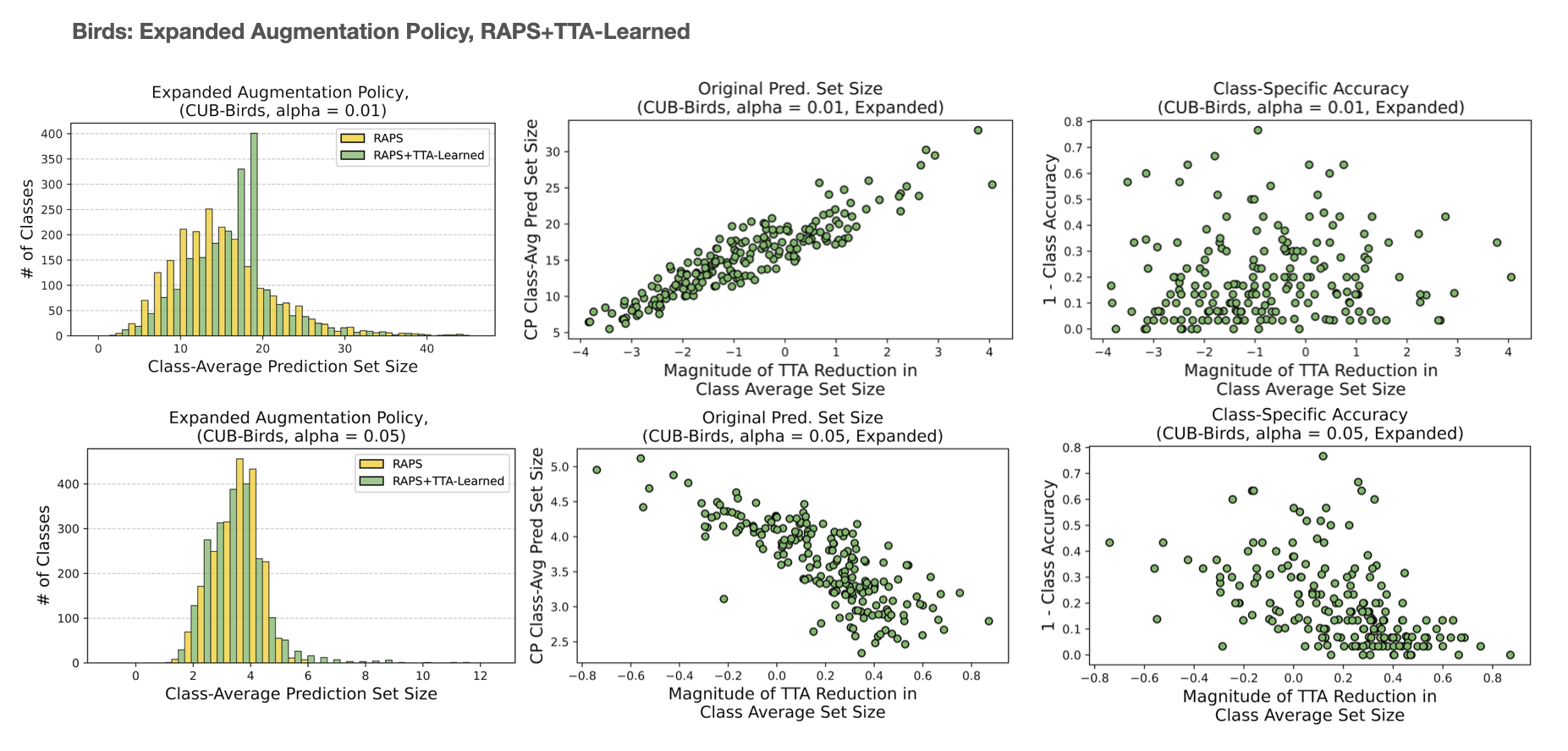}
\end{center}
\vspace{-1em}
\caption{\textbf{Class-specific performance for CUB-Birds}, for $\alpha = .01$ (top) and $\alpha = .05\%$ (bottom). These graphs show an example for which TTA-Learned does \emph{not} produce improvements in average prediction set size (computed across all examples). Interestingly, behavior on a class-specific level is different between $\alpha = .01$ and $\alpha = .05$. For $\alpha = .01$, results are consistent with other datasets: classes which originally receive large prediction set sizes and classes which are more difficult benefit most from the addition of TTA. For $\alpha = .05$, the exact opposite is true. While a majority of classes are hurt by TTA, classes that benefit from TTA are easier and receive smaller prediction set sizes.}
\label{fig:class_specific_birds_expanded}\vspace{-.5em}\vspace{-.5em}
\end{figure*}

We replicate the class-specific analysis for ImageNet at a value of $\alpha = .05$ (Figure \ref{fig:class_specific_imagenet_.05_expanded}), iNaturalist (Figure \ref{fig:class_specific_inat_expanded}),  and CUB-Birds (Figure \ref{fig:class_specific_birds_expanded}). All trends are consistent with results in the main text, save for one notable exception: when TTA-Learned is applied to CUB-Birds, prediction set sizes of the classes with the \emph{smallest} prediction set sizes and classes that are \emph{easier} to predict benefit most from TTA. The significance of the relationship between original prediction set size and TTA improvement disappears when conducted on an example level in this setting. This could be a result of class imbalance in the dataset; it is possible that the class-average prediction set size obscures important variation in CUB-Birds.

\subsection{Impact of augmentation policy size}

\begin{figure*}
\begin{center}
\includegraphics[width=.9\textwidth]{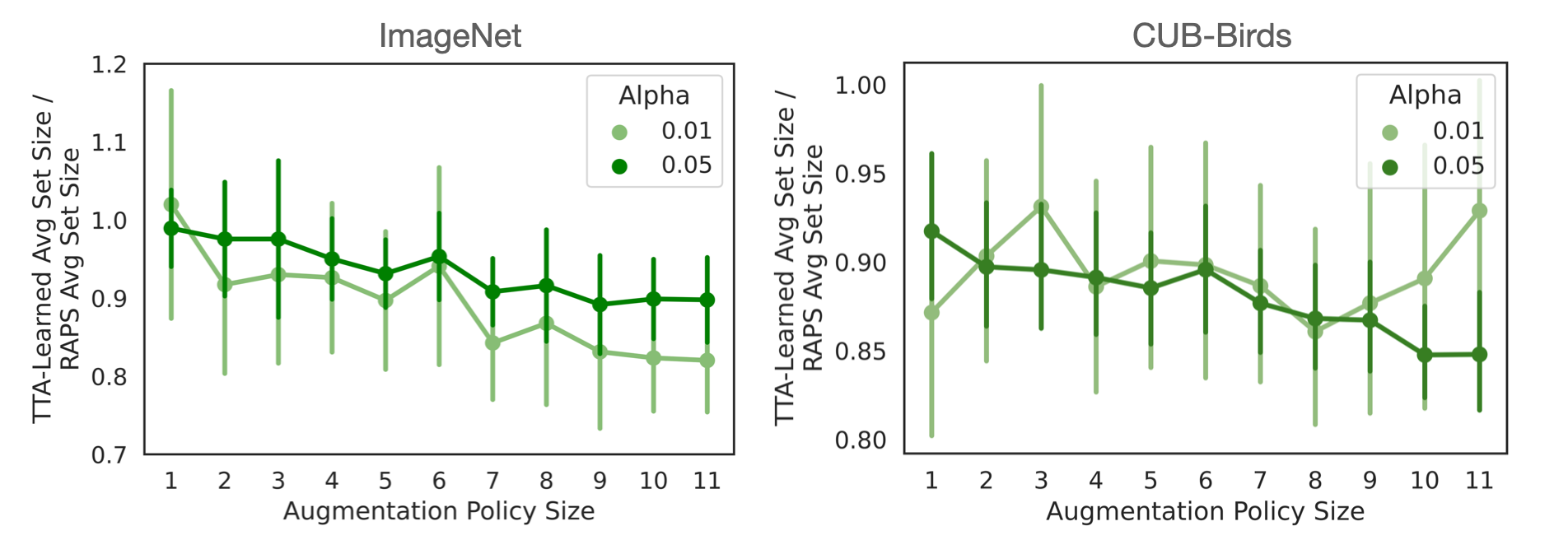}
\end{center}
\vspace{-1em}
\caption{\textbf{Impact of augmentation policy size.} We see that larger policy sizes translate to a greater improvement (in terms of the ratio of average prediction set sizes using RAPS+TTA-Learned  to average prediction set sizes using RAPS alone) for $\alpha = .05$. For $\alpha = .01$, there is no clear trend.}
\label{fig:aug_policy}
\vspace{-.5em}
\end{figure*}

We also analyze the impact of augmentation policy size on  average prediction set size for CUB-Birds (Figure \ref{fig:aug_policy}), to understand  if additional augmentations may produce larger reductions in  set size than we observe. Larger augmentation policies appear to provide an improvement to average prediction set size at $\alpha = .05$, but offer little improvement  for $\alpha = .01$.

\subsection{Impact of TTA data split}
\label{supp:tta_data}

\begin{figure*}
\begin{center}
\includegraphics[width=.45\textwidth]{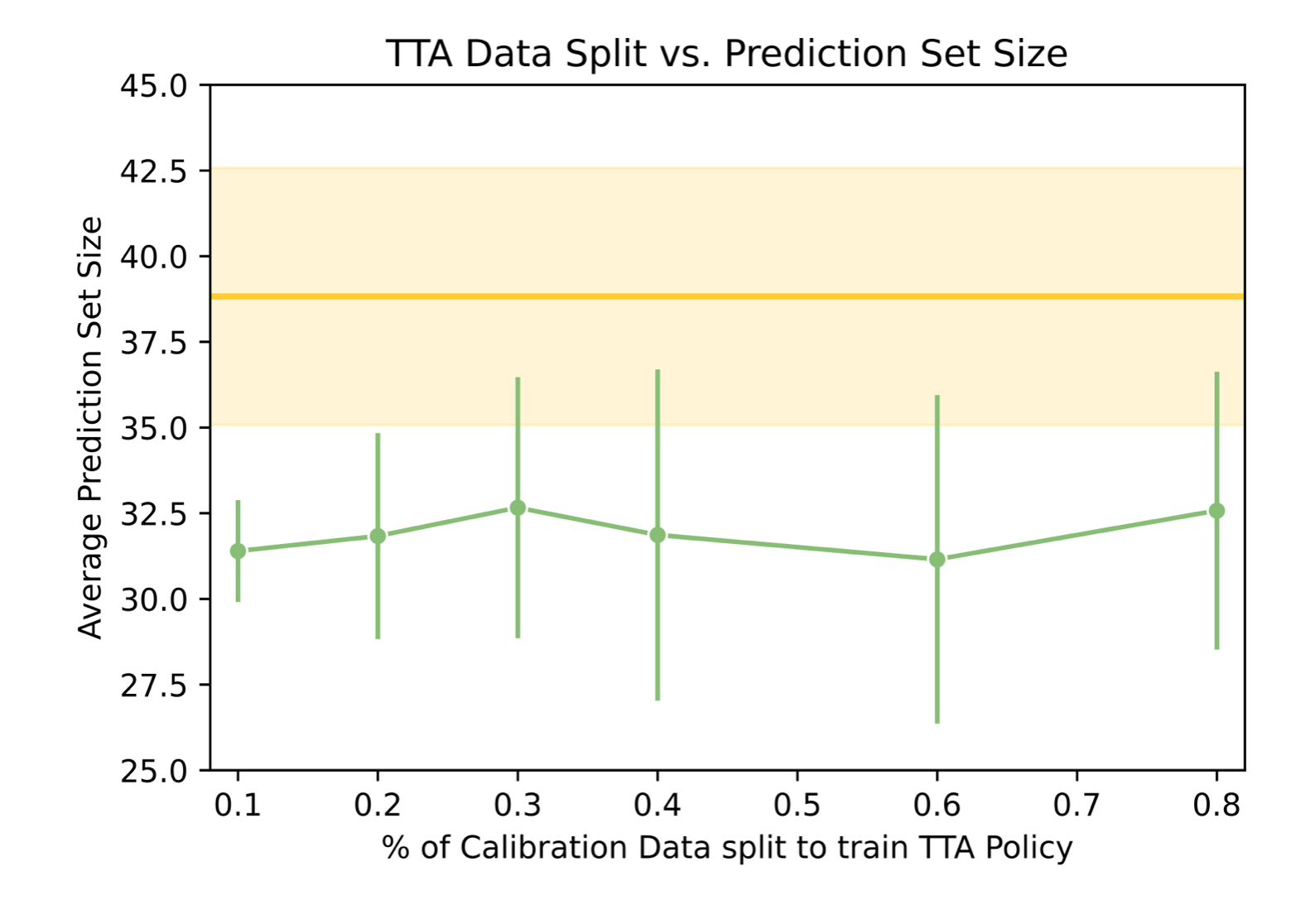}
\end{center}
\vspace{-1em}
\caption{\textbf{Robustness to size of dataset used to train test-time augmentation policy.} We plot the percentage of data used to train the TTA policy on the x-axis and the average prediction set size on the y-axis. Error bars describe variance over 10 random splits of the calibration and test set. We can make two observations: 1) as the data used to train the  TTA policy increases and the data used to estimate the conformal threshold decreases, variance in performance grows and 2) across a wide range of  data splits, learned TTA policies (green) introduce improvements to achieved prediction set sizes compared to the original probabilities (gold). These results also suggest that relatively little training data is required to  learn a useful test-time augmentation policy; in this  case, 2-3 images per class, or 10\% of the available labeled data.}
\label{fig:supp_tta_data}
\vspace{-.5em}
\end{figure*}

Learning the test-time augmentation policy requires a set of labeled data \emph{distinct} from those used to select the conformal threshold. This introduces a trade-off: more labeled data for test-time augmentation may result in more accurate weights, but a less accurate conformal threshold, and vice versa. We study this tradeoff empirically in the context of ImageNet and the expanded augmentation policy and show results in Figure \ref{fig:supp_tta_data}. We find that, as more data is taken away from the conformal calibration set, variance in performance grows. This is in line with our intuition; we have fewer examples to approximate the distribution of conformal scores.  However, at all percentages, test-time augmentation introduces a significant improvement in prediction set sizes over using all the labeled examples, and their original probabilities, to determine the threshold. This suggests that the benefits TTA confers outweigh the costs to the estimation of the conformal threshold, a practically useful insight to those who wish to apply conformal prediction in practice6

\subsection{Impact of calibration set size}
\label{supp:cal_set_size}

\begin{figure*}
\begin{center}
\includegraphics[width=\textwidth]{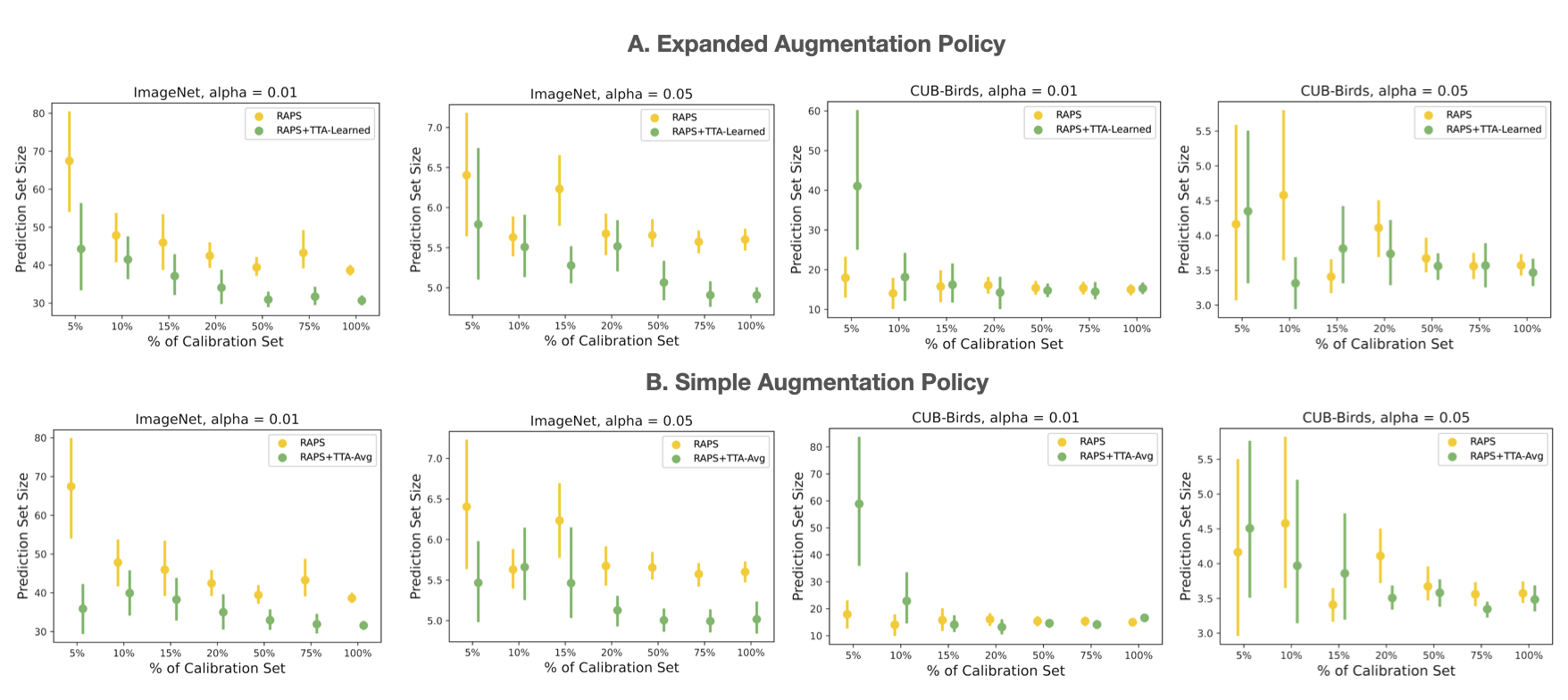}
\end{center}
\vspace{-1em}
\caption{\textbf{Impact of calibration set size.} We plot the relationship between calibration set size and average prediction set size across two values of alpha, two augmentation policies, and two  datasets (ImageNet and CUB-Birds). For ImageNet, larger calibration set sizes correlate with larger and more consistent improvements from the addition of TTA, where the improvement flattens out for calibration set sizes larger than 50\%, or 12,500 images  (12-13 per class). TTA does appear to be able to improve average prediction set size even with a calibration set size of 1,250 (5\% of original ImageNet calibration set size). For CUB-Birds, a dataset on which TTA does not perform as well, we see that TTA performs comparably to RAPS alone  the larger the calibration set.}
\label{fig:cal_set_size}
\vspace{-.5em}
\end{figure*}

We plot the relationship between calibration set size and average prediction set size in Figure \ref{fig:cal_set_size} across two augmentation policies, two datasets, and two values of $\alpha$. We see that TTA is more  effective the larger the calibration set, in the context of ImageNet. In the context of CUB-Birds, it appears that TTA approaches equivalence with the conformal score alone as the calibration set size increases.

\subsection{Impact of different backbone architecture}

Our results in the main text are limited to a single architecture (residual networks). Here, we provide evidence of generalizability to different architectures by replicating our ImageNet results using MobileNetV2, across a range of coverage guarantees and both augmentation policies (Table \ref{tab:mobilenetv2}) and find consistent results, which support the versatility of the proposed method.

\renewcommand{\arraystretch}{1.3}

\begin{table*}[h]

\resizebox{\textwidth}{!}{%
\rowcolors{2}{gray!20}{white}

\begin{tabular}{cl|c|c|c|c|c|c}
\toprule
  &  & \multicolumn{3}{c|}{Expanded Aug Policy} & \multicolumn{3}{c}{Simple Aug Policy} \\
\toprule
 Alpha & Method & ImageNet & iNaturalist & CUB-Birds & ImageNet & iNaturalist & CUB-Birds \\
\toprule
\hline
0.01 & APS & 98.493 $\pm$ 3.075 & 131.681 $\pm$ 3.515 & 19.436 $\pm$ 0.995 & 98.493 $\pm$ 3.075 & \textbf{131.681 $\pm$ 3.515} & \textbf{19.436 $\pm$ 0.995}\\

0.01 & APS+TTA-Avg & \textbf{68.714 $\pm$ 2.856} & \textbf{84.546 $\pm$ 3.655} & \textbf{17.715 $\pm$ 1.523} & \textbf{92.027 $\pm$ 4.797} & 145.401 $\pm$ 4.635 & \textbf{19.152 $\pm$ 1.667}\\

0.01 & APS+TTA-Learned & \textbf{69.009 $\pm$ 2.156} & \textbf{85.093 $\pm$ 2.768} & \textbf{17.766 $\pm$ 1.608} & \textbf{90.613 $\pm$ 6.421} & 144.134 $\pm$ 4.371 & \textbf{18.552 $\pm$ 1.326}\\

\hline
0.05 & APS & 19.820 $\pm$ 0.482 & 33.481 $\pm$ 0.786 & 5.921 $\pm$ 0.192 & 19.820 $\pm$ 0.482 & \textbf{33.481 $\pm$ 0.786} & \textbf{5.921 $\pm$ 0.192}\\

0.05 & APS+TTA-Avg & 14.308 $\pm$ 0.279 & \textbf{26.021 $\pm$ 0.282} & \textbf{4.870 $\pm$ 0.208} & \textbf{18.862 $\pm$ 0.498} & 37.370 $\pm$ 0.735 & 6.306 $\pm$ 0.350\\

0.05 & APS+TTA-Learned & \textbf{14.084 $\pm$ 0.241} & \textbf{26.289 $\pm$ 0.529} & \textbf{4.913 $\pm$ 0.145} & \textbf{19.119 $\pm$ 0.479} & 36.940 $\pm$ 0.632 & 6.361 $\pm$ 0.480\\

\hline
0.10 & APS & 8.969 $\pm$ 0.158 & 16.755 $\pm$ 0.394 & 3.455 $\pm$ 0.164 & 8.969 $\pm$ 0.158 & \textbf{16.755 $\pm$ 0.394} & \textbf{3.455 $\pm$ 0.164}\\

0.10 & APS+TTA-Avg & \textbf{7.193 $\pm$ 0.101} & \textbf{14.583 $\pm$ 0.333} & \textbf{3.108 $\pm$ 0.114} & \textbf{8.787 $\pm$ 0.136} & 18.300 $\pm$ 0.418 & \textbf{3.609 $\pm$ 0.135}\\

0.10 & APS+TTA-Learned & \textbf{7.215 $\pm$ 0.106} & \textbf{14.538 $\pm$ 0.395} & \textbf{3.046 $\pm$ 0.073} & \textbf{8.813 $\pm$ 0.180} & 18.086 $\pm$ 0.420 & 3.638 $\pm$ 0.146\\

\hline
\end{tabular}
}
\caption{We replicate our experiments across coverage levels and datasets using APS, another conformal score. TTA-Learned combined with the expanded augmentation policy produces the smallest set sizes across all comparisons. Interestingly, the simple augmentation policy is not as effective in the context of iNaturalist when using APS. }
\label{results-aps}
\vspace{.5em}

\resizebox{\textwidth}{!}{%
\rowcolors{2}{gray!20}{white}

\begin{tabular}{cl|c|c|c|c|c|c}
\toprule
  &  & \multicolumn{3}{c|}{Expanded Aug Policy} & \multicolumn{3}{c}{Simple Aug Policy} \\
\toprule
 Alpha & Method & ImageNet & iNaturalist & CUB-Birds & ImageNet & iNaturalist & CUB-Birds \\
\toprule
\hline
0.01 & APS & 0.980 $\pm$ 0.001 & 0.986 $\pm$ 0.000 & 0.985 $\pm$ 0.001 & 0.980 $\pm$ 0.001 & \textbf{0.986 $\pm$ 0.000} & \textbf{0.985 $\pm$ 0.001}\\

0.01 & APS+TTA-Avg & \textbf{0.985 $\pm$ 0.001} & \textbf{0.989 $\pm$ 0.001} & \textbf{0.989 $\pm$ 0.002} & \textbf{0.981 $\pm$ 0.001} & 0.987 $\pm$ 0.000 & \textbf{0.986 $\pm$ 0.003}\\

0.01 & APS+TTA-Learned & \textbf{0.985 $\pm$ 0.001} & \textbf{0.989 $\pm$ 0.001} & \textbf{0.990 $\pm$ 0.002} & \textbf{0.980 $\pm$ 0.002} & 0.987 $\pm$ 0.000 & \textbf{0.985 $\pm$ 0.002}\\

\hline
0.05 & APS & 0.931 $\pm$ 0.002 & 0.952 $\pm$ 0.001 & 0.945 $\pm$ 0.004 & 0.931 $\pm$ 0.002 & \textbf{0.952 $\pm$ 0.001} & \textbf{0.945 $\pm$ 0.004}\\

0.05 & APS+TTA-Avg & 0.944 $\pm$ 0.002 & \textbf{0.956 $\pm$ 0.001} & \textbf{0.949 $\pm$ 0.005} & \textbf{0.937 $\pm$ 0.002} & 0.960 $\pm$ 0.001 & 0.949 $\pm$ 0.004\\

0.05 & APS+TTA-Learned & \textbf{0.943 $\pm$ 0.002} & \textbf{0.957 $\pm$ 0.001} & \textbf{0.950 $\pm$ 0.005} & \textbf{0.937 $\pm$ 0.002} & 0.959 $\pm$ 0.001 & 0.950 $\pm$ 0.005\\

\hline
0.10 & APS & 0.896 $\pm$ 0.002 & 0.923 $\pm$ 0.001 & 0.915 $\pm$ 0.006 & 0.896 $\pm$ 0.002 & \textbf{0.923 $\pm$ 0.001} & \textbf{0.915 $\pm$ 0.006}\\

0.10 & APS+TTA-Avg & \textbf{0.903 $\pm$ 0.002} & \textbf{0.930 $\pm$ 0.001} & \textbf{0.920 $\pm$ 0.007} & \textbf{0.905 $\pm$ 0.002} & 0.933 $\pm$ 0.001 & \textbf{0.922 $\pm$ 0.005}\\

0.10 & APS+TTA-Learned & \textbf{0.904 $\pm$ 0.002} & \textbf{0.930 $\pm$ 0.001} & \textbf{0.918 $\pm$ 0.006} & \textbf{0.906 $\pm$ 0.002} & 0.932 $\pm$ 0.001 & 0.922 $\pm$ 0.004\\

\hline
\end{tabular}

}
\caption{Coverage values associated with experiments in Table \ref{results-aps}. TTA-Learned produces significant improvements in coverage --- larger in magnitude than in conjunction with RAPS ---  across when using the expanded augmentation policy. TTA-Learned produces no drops in coverage when using the simple augmentation policy, a nd produces improvements at $\alpha = .01$ and $\alpha = .05$.}
\label{results-aps-coverage}

\end{table*}

\renewcommand{\arraystretch}{1.3}

\begin{table*}[ht]
\centering

\resizebox{\textwidth}{!}{%
\rowcolors{2}{gray!20}{white}
\begin{tabular}{cl|c|c|c|c|c|c}
\toprule
  &  & \multicolumn{3}{c|}{Expanded Aug Policy} & \multicolumn{3}{c}{Simple Aug Policy} \\
\toprule
 Alpha & Method & ResNet-50 & ResNet-101 & ResNet-152 &  ResNet-50  & ResNet-101 & ResNet-152 \\
\toprule
\hline
0.01 & APS & 98.493 $\pm$ 3.075 & 88.279 $\pm$ 4.121 & 79.231 $\pm$ 4.570 & 98.493 $\pm$ 3.075 & 88.279 $\pm$ 4.121 & 79.231 $\pm$ 4.570\\

0.01 & APS+TTA-Avg & \textbf{68.714 $\pm$ 2.856} & \textbf{64.197 $\pm$ 2.336} & \textbf{62.885 $\pm$ 3.125} & \textbf{92.027 $\pm$ 4.797} & \textbf{77.344 $\pm$ 2.214} & \textbf{73.377 $\pm$ 3.600}\\

0.01 & APS+TTA-Learned & \textbf{69.009 $\pm$ 2.156} & \textbf{64.852 $\pm$ 2.823} & \textbf{64.045 $\pm$ 3.398} & \textbf{90.613 $\pm$ 6.421} & \textbf{78.627 $\pm$ 4.101} & \textbf{74.571 $\pm$ 3.516}\\

\hline
0.05 & APS & 19.820 $\pm$ 0.482 & 15.830 $\pm$ 0.611 & 14.437 $\pm$ 0.591 & 19.820 $\pm$ 0.482 & 15.830 $\pm$ 0.611 & \textbf{14.437 $\pm$ 0.591}\\

0.05 & APS+TTA-Avg & 14.308 $\pm$ 0.279 & \textbf{11.085 $\pm$ 0.267} & \textbf{10.605 $\pm$ 0.373} & \textbf{18.862 $\pm$ 0.498} & \textbf{15.039 $\pm$ 0.405} & \textbf{14.206 $\pm$ 0.499}\\

0.05 & APS+TTA-Learned & \textbf{14.084 $\pm$ 0.241} & \textbf{11.118 $\pm$ 0.209} & \textbf{10.595 $\pm$ 0.368} & \textbf{19.119 $\pm$ 0.479} & \textbf{15.011 $\pm$ 0.346} & \textbf{14.252 $\pm$ 0.486}\\

\hline
0.10 & APS & 8.969 $\pm$ 0.158 & 6.671 $\pm$ 0.175 & 6.134 $\pm$ 0.163 & 8.969 $\pm$ 0.158 & \textbf{6.671 $\pm$ 0.175} & \textbf{6.134 $\pm$ 0.163}\\

0.10 & APS+TTA-Avg & \textbf{7.193 $\pm$ 0.101} & \textbf{5.454 $\pm$ 0.098} & \textbf{5.111 $\pm$ 0.096} & \textbf{8.787 $\pm$ 0.136} & 6.838 $\pm$ 0.143 & 6.309 $\pm$ 0.178\\

0.10 & APS+TTA-Learned & \textbf{7.215 $\pm$ 0.106} & \textbf{5.490 $\pm$ 0.090} & \textbf{5.131 $\pm$ 0.061} & \textbf{8.813 $\pm$ 0.180} & 6.826 $\pm$ 0.121 & 6.311 $\pm$ 0.123\\

\hline
\end{tabular}
}
\caption{Results across base classifiers using APS alone, APS + TTA-Avg, and APS + TTA-learned in conjunction with the expanded augmentation policy (left) and simple augmentation policy (right). TTA-Learned and the expanded augmentation policy produce the smallest prediction sets (on average). }
\label{model-results-aps}

\vspace{1em}
\resizebox{\textwidth}{!}{%
\rowcolors{2}{gray!20}{white}
\begin{tabular}{cl|c|c|c|c|c|c}
\toprule
  &  & \multicolumn{3}{c|}{Expanded Aug Policy} & \multicolumn{3}{c}{Simple Aug Policy} \\
\toprule
 Alpha & Method & ResNet-50 & ResNet-101 & ResNet-152 &  ResNet-50  & ResNet-101 & ResNet-152 \\
\toprule
\hline
0.01 & APS & 0.980 $\pm$ 0.001 & 0.979 $\pm$ 0.002 & 0.978 $\pm$ 0.002 & \textbf{0.980 $\pm$ 0.001} & \textbf{0.979 $\pm$ 0.002} & \textbf{0.978 $\pm$ 0.002}\\

0.01 & APS+TTA-Avg & \textbf{0.985 $\pm$ 0.001} & \textbf{0.985 $\pm$ 0.001} & \textbf{0.984 $\pm$ 0.001} & \textbf{0.981 $\pm$ 0.001} & \textbf{0.980 $\pm$ 0.001} & \textbf{0.978 $\pm$ 0.002}\\

0.01 & APS+TTA-Learned & \textbf{0.985 $\pm$ 0.001} & \textbf{0.985 $\pm$ 0.001} & \textbf{0.984 $\pm$ 0.001} & \textbf{0.980 $\pm$ 0.002} & \textbf{0.980 $\pm$ 0.002} & \textbf{0.979 $\pm$ 0.002}\\

0.05 & APS & 0.931 $\pm$ 0.002 & 0.930 $\pm$ 0.002 & 0.929 $\pm$ 0.002 & 0.931 $\pm$ 0.002 & 0.930 $\pm$ 0.002 & 0.929 $\pm$ 0.002\\

0.05 & APS+TTA-Avg & \textbf{0.944 $\pm$ 0.002} & \textbf{0.942 $\pm$ 0.001} & \textbf{0.942 $\pm$ 0.002} & \textbf{0.937 $\pm$ 0.002} & \textbf{0.935 $\pm$ 0.002} & \textbf{0.934 $\pm$ 0.002}\\

0.05 & APS+TTA-Learned & 0.943 $\pm$ 0.002 & \textbf{0.942 $\pm$ 0.001} & \textbf{0.942 $\pm$ 0.002} & \textbf{0.937 $\pm$ 0.002} & \textbf{0.935 $\pm$ 0.001} & \textbf{0.934 $\pm$ 0.002}\\

0.10 & APS & 0.896 $\pm$ 0.002 & 0.892 $\pm$ 0.002 & 0.893 $\pm$ 0.002 & 0.896 $\pm$ 0.002 & 0.892 $\pm$ 0.002 & 0.893 $\pm$ 0.002\\

0.10 & APS+TTA-Avg & \textbf{0.903 $\pm$ 0.002} & \textbf{0.901 $\pm$ 0.001} & \textbf{0.902 $\pm$ 0.001} & \textbf{0.905 $\pm$ 0.002} & \textbf{0.903 $\pm$ 0.001} & \textbf{0.903 $\pm$ 0.002}\\

0.10 & APS+TTA-Learned & \textbf{0.904 $\pm$ 0.002} & \textbf{0.902 $\pm$ 0.001} & \textbf{0.902 $\pm$ 0.001} & \textbf{0.906 $\pm$ 0.002} & \textbf{0.903 $\pm$ 0.002} & \textbf{0.903 $\pm$ 0.002}\\
\hline

\end{tabular}
}
\label{model-results-aps-coverage}
\caption{Coverage values for APS and TTA variants of APS across base classifiers, using ImageNet. TTA-Learned or TTA-Avg in combination with the expanded augmentation policy significantly improve coverage in every comparison.}

\end{table*}

\renewcommand{\arraystretch}{1.2}

\begin{table*}
\centering

\resizebox{.8\textwidth}{!}{%
\rowcolors{2}{gray!20}{white}
\begin{tabular}{l|ccc|ccc}
\toprule
  &  \multicolumn{3}{c}{Expanded Aug Policy} & \multicolumn{3}{|c}{Simple Aug Policy} \\
\toprule
Method & ResNet50 & ResNet101 & ResNet152 & ResNet50 & ResNet101 & ResNet152 \\
\midrule
Original & 0.761 ± 0.002 & 0.773 ± 0.001 & 0.783 ± 0.002 & 0.761 ± 0.002 & 0.773 ± 0.001 & 0.783 ± 0.002 \\
TTA-Avg & 0.764 ± 0.002 & 0.778 ± 0.001 & 0.788 ± 0.002 & 0.77 ± 0.002 & 0.783 ± 0.001 & 0.792 ± 0.002 \\
TTA-Learned & 0.771 ± 0.002 & 0.785 ± 0.001 & 0.793 ± 0.002 & 0.771 ± 0.002 & 0.784 ± 0.001 & 0.793 ± 0.002 \\
\bottomrule
\end{tabular}
}
\caption{\textbf{TTA effect on classifier performance.} We report differences in classifier performance using a learned test-time augmentation policy compared to a simple average (TTA-Avg) and no test-time augmentation (Original). TTA-Learned offers small improvements over a simpler average and the original model across architectures. FILL IN THE REST, explain how TTA's improvement to Top-1 accuracy alone is small, and does not fully explain the value of test-time augmentation to conformal prediction.  }
\label{model-top1-acc-improvements}
\end{table*}

\renewcommand{\arraystretch}{1.3}

\begin{table*}[htp]
\centering
\begin{minipage}[t]{\textwidth}
\resizebox{1\textwidth}{!}{%
\rowcolors{2}{gray!20}{white}
\begin{tabular}{cl|cc|cc|cc}
\toprule
 & & \multicolumn{2}{c|}{ImageNet} & \multicolumn{2}{c|}{iNaturalist} & \multicolumn{2}{c}{CUB-Birds}\\
\hline
Alpha & Method & Prediction Set Size & Empirical Coverage & Prediction Set Size & Empirical Coverage & Prediction Set Size & Empirical Coverage\\
\hline
0.01 & Top-1 & 1.000 $\pm$ 0.000 & 0.761 $\pm$ 0.002 & 1.000 $\pm$ 0.000 & 0.766 $\pm$ 0.001 & 1.000 $\pm$ 0.000 & 0.804 $\pm$ 0.008\\

0.01 & Top-5 & 5.000 $\pm$ 0.000 & 0.928 $\pm$ 0.001 & 5.000 $\pm$ 0.000 & 0.915 $\pm$ 0.001 & 5.000 $\pm$ 0.000 & 0.959 $\pm$ 0.003\\

0.01 & RAPS & 37.751 $\pm$ 2.334 & 0.990 $\pm$ 0.001 & 61.437 $\pm$ 6.067 & 0.990 $\pm$ 0.001 & 15.293 $\pm$ 2.071 & 0.990 $\pm$ 0.001\\

0.01 & RAPS+TTA-Avg & 35.600 $\pm$ 2.200 & 0.991 $\pm$ 0.001 & 57.073 $\pm$ 5.914 & 0.990 $\pm$ 0.001 & 13.111 $\pm$ 2.470 & 0.991 $\pm$ 0.002\\

0.01 & RAPS+TTA-Learned & 31.248 $\pm$ 2.177 & 0.990 $\pm$ 0.001 & 53.195 $\pm$ 4.884 & 0.990 $\pm$ 0.001 & 14.045 $\pm$ 1.323 & 0.991 $\pm$ 0.002\\

\hline
0.05 & Top-1 & 1.000 $\pm$ 0.000 & 0.761 $\pm$ 0.002 & 1.000 $\pm$ 0.000 & 0.766 $\pm$ 0.001 & 1.000 $\pm$ 0.000 & 0.804 $\pm$ 0.008\\

0.05 & Top-5 & 5.000 $\pm$ 0.000 & 0.928 $\pm$ 0.001 & 5.000 $\pm$ 0.000 & 0.915 $\pm$ 0.001 & 5.000 $\pm$ 0.000 & 0.959 $\pm$ 0.003\\

0.05 & RAPS & 5.637 $\pm$ 0.357 & 0.951 $\pm$ 0.002 & 7.991 $\pm$ 1.521 & 0.954 $\pm$ 0.002 & 3.624 $\pm$ 0.361 & 0.955 $\pm$ 0.007\\

0.05 & RAPS+TTA-Avg & 5.318 $\pm$ 0.113 & 0.951 $\pm$ 0.001 & 7.067 $\pm$ 0.344 & 0.952 $\pm$ 0.002 & 3.116 $\pm$ 0.210 & 0.954 $\pm$ 0.007\\

0.05 & RAPS+TTA-Learned & 4.889 $\pm$ 0.168 & 0.952 $\pm$ 0.001 & 6.682 $\pm$ 0.447 & 0.954 $\pm$ 0.002 & 3.571 $\pm$ 0.576 & 0.957 $\pm$ 0.007\\

\hline
0.10 & Top-1 & 1.000 $\pm$ 0.000 & 0.761 $\pm$ 0.002 & 1.000 $\pm$ 0.000 & 0.766 $\pm$ 0.001 & 1.000 $\pm$ 0.000 & 0.804 $\pm$ 0.008\\

0.10 & Top-5 & 5.000 $\pm$ 0.000 & 0.928 $\pm$ 0.001 & 5.000 $\pm$ 0.000 & 0.915 $\pm$ 0.001 & 5.000 $\pm$ 0.000 & 0.959 $\pm$ 0.003\\

0.10 & RAPS & 2.548 $\pm$ 0.074 & 0.906 $\pm$ 0.004 & 2.914 $\pm$ 0.116 & 0.907 $\pm$ 0.003 & 2.038 $\pm$ 0.153 & 0.919 $\pm$ 0.014\\

0.10 & RAPS+TTA-Avg & 2.470 $\pm$ 0.071 & 0.905 $\pm$ 0.005 & 2.740 $\pm$ 0.026 & 0.908 $\pm$ 0.002 & 1.780 $\pm$ 0.139 & 0.912 $\pm$ 0.014\\

0.10 & RAPS+TTA-Learned & 2.312 $\pm$ 0.054 & 0.905 $\pm$ 0.004 & 2.625 $\pm$ 0.043 & 0.909 $\pm$ 0.003 & 1.893 $\pm$ 0.187 & 0.919 $\pm$ 0.016\\

\hline
\end{tabular}

}
\caption{\textbf{Comparison to Top-1 and Top-5 baselines.} Results comparing performance against Top-K baselines. In each setting, conformal prediction produces either smaller set sizes, higher coverage, or both compared to the Top-K baselines.}
\label{results-topk}
\end{minipage}
\end{table*}

\renewcommand{\arraystretch}{1.2}

\begin{table*}[htp]
\begin{minipage}[t]{\textwidth}
\centering
\resizebox{.8\textwidth}{!}{%
\rowcolors{2}{gray!20}{white}
\begin{tabular}{cl|c|c|c|c|c|c}
\toprule
  &  & \multicolumn{2}{c|}{Expanded Aug Policy} & \multicolumn{2}{c}{Simple Aug Policy} \\
\toprule
 Alpha & Method & ImageNet  & CUB-Birds & ImageNet & CUB-Birds \\
\toprule
\hline
0.01 & RAPS+TTA-Learned+Focal & 32.612 $\pm$ 3.799 & 13.416 $\pm$ 1.991 & 31.230 $\pm$ 1.510 & 15.503 $\pm$ 2.364\\

0.01 & RAPS+TTA-Learned+Conformal & 32.257 $\pm$ 3.608 & 13.776 $\pm$ 2.198 & 31.716 $\pm$ 2.078 & 14.432 $\pm$ 2.184\\

0.01 & RAPS+TTA-Learned+CE & 31.248 $\pm$ 2.177 & 14.045 $\pm$ 1.323 & 32.702 $\pm$ 2.409 & 13.803 $\pm$ 1.734\\

\hline
0.05 & RAPS+TTA-Learned+Focal & 4.906 $\pm$ 0.195 & 3.194 $\pm$ 0.202 & 4.956 $\pm$ 0.239 & 3.313 $\pm$ 0.331\\

0.05 & RAPS+TTA-Learned+Conformal & 4.867 $\pm$ 0.122 & 3.302 $\pm$ 0.312 & 4.996 $\pm$ 0.405 & 3.412 $\pm$ 0.406\\

0.05 & RAPS+TTA-Learned+CE & 4.889 $\pm$ 0.168 & 3.571 $\pm$ 0.576 & 5.040 $\pm$ 0.176 & 3.290 $\pm$ 0.186\\

\hline
0.10 & RAPS+TTA-Learned+Focal & 2.363 $\pm$ 0.085 & 1.791 $\pm$ 0.102 & 2.308 $\pm$ 0.045 & 1.860 $\pm$ 0.131\\

0.10 & RAPS+TTA-Learned+Conformal & 2.308 $\pm$ 0.068 & 1.865 $\pm$ 0.163 & 2.330 $\pm$ 0.072 & 1.868 $\pm$ 0.122\\

0.10 & RAPS+TTA-Learned+CE & 2.312 $\pm$ 0.054 & 1.893 $\pm$ 0.187 & 2.362 $\pm$ 0.065 & 1.840 $\pm$ 0.106\\

\hline
\end{tabular}

}
\caption{\textbf{Alternate training objectives.} Results across datasets for two augmentation policies and three coverage specifications using a focal loss. We set $\gamma$ to be 1, in line with prior work \citep{einbinder_training_nodate}. Each entry corresponds to the average prediction set size across 10 calibration/test splits. Both the focal and conformal loss do not outperform the cross-entropy loss; for simplicity, we report all results using the cross-entropy loss.}
\label{tab:alternate-losses}
\end{minipage}
\end{table*}

\renewcommand{\arraystretch}{1.3}

\begin{table*}[htp]
\centering
\begin{minipage}[t]{\textwidth}
\resizebox{1\textwidth}{!}{%
\rowcolors{2}{gray!20}{white}
\begin{tabular}{cl|c|c|c|c|c|c}
\toprule
  &  & \multicolumn{3}{c|}{Expanded Aug Policy} & \multicolumn{3}{c}{Simple Aug Policy} \\
\toprule
 Alpha & Method & ImageNet & iNaturalist & CUB-Birds & ImageNet & iNaturalist & CUB-Birds \\
\hline
0.01 & RAPS & \textbf{0.990 $\pm$ 0.001} & \textbf{0.990 $\pm$ 0.001} & \textbf{0.990 $\pm$ 0.001} & \textbf{0.990 $\pm$ 0.001} & \textbf{0.990 $\pm$ 0.001} & \textbf{0.990 $\pm$ 0.001}\\

0.01 & RAPS+TTA-Avg & \textbf{0.991 $\pm$ 0.001} & \textbf{0.990 $\pm$ 0.001} & \textbf{0.991 $\pm$ 0.002} & \textbf{0.990 $\pm$ 0.001} & \textbf{0.990 $\pm$ 0.001} & \textbf{0.991 $\pm$ 0.002}\\

0.01 & RAPS+TTA-Learned & \textbf{0.990 $\pm$ 0.001} & \textbf{0.990 $\pm$ 0.001} & \textbf{0.991 $\pm$ 0.002} & \textbf{0.990 $\pm$ 0.001} & \textbf{0.990 $\pm$ 0.001} & \textbf{0.990 $\pm$ 0.002}\\

\hline
0.05 & RAPS & \textbf{0.951 $\pm$ 0.002} & \textbf{0.954 $\pm$ 0.002} & \textbf{0.955 $\pm$ 0.007} & \textbf{0.951 $\pm$ 0.002} & \textbf{0.954 $\pm$ 0.002} & \textbf{0.955 $\pm$ 0.007}\\

0.05 & RAPS+TTA-Avg & \textbf{0.951 $\pm$ 0.001} & \textbf{0.952 $\pm$ 0.002} & \textbf{0.954 $\pm$ 0.007} & \textbf{0.951 $\pm$ 0.001} & \textbf{0.953 $\pm$ 0.003} & \textbf{0.957 $\pm$ 0.004}\\

0.05 & RAPS+TTA-Learned & \textbf{0.952 $\pm$ 0.001} & \textbf{0.954 $\pm$ 0.002} & \textbf{0.957 $\pm$ 0.007} & \textbf{0.951 $\pm$ 0.002} & \textbf{0.952 $\pm$ 0.002} & \textbf{0.956 $\pm$ 0.007}\\

\hline
0.10 & RAPS & \textbf{0.906 $\pm$ 0.004} & \textbf{0.907 $\pm$ 0.003} & \textbf{0.919 $\pm$ 0.014} & \textbf{0.906 $\pm$ 0.004} & \textbf{0.907 $\pm$ 0.003} & \textbf{0.919 $\pm$ 0.014}\\

0.10 & RAPS+TTA-Avg & \textbf{0.905 $\pm$ 0.005} & \textbf{0.908 $\pm$ 0.002} & \textbf{0.912 $\pm$ 0.014} & \textbf{0.905 $\pm$ 0.004} & \textbf{0.908 $\pm$ 0.002} & \textbf{0.915 $\pm$ 0.010}\\

0.10 & RAPS+TTA-Learned & \textbf{0.905 $\pm$ 0.004} & \textbf{0.909 $\pm$ 0.003} & \textbf{0.919 $\pm$ 0.016} & \textbf{0.907 $\pm$ 0.004} & \textbf{0.908 $\pm$ 0.003} & \textbf{0.913 $\pm$ 0.011}\\

\hline
\end{tabular}

}
\caption{\textbf{Comparison of achieved coverage.} Coverage values for RAPS, RAPS+TTA-Avg, and RAPS+TTA-Learned across datasets and coverage values. RAPS+TTA-Learned never decreases the coverage achieved by RAPS alone, and in some cases, improves it significantly (as in the case of ImageNet and iNaturalist).}
\label{results-raps-coverage}
\end{minipage}
\end{table*}

\renewcommand{\arraystretch}{1.3}

\begin{table*}[ht]
\centering

\resizebox{\textwidth}{!}{%
\rowcolors{2}{gray!20}{white}
\begin{tabular}{cl|c|c|c|c|c|c}
\toprule
  &  & \multicolumn{3}{c|}{Expanded Aug Policy} & \multicolumn{3}{c}{Simple Aug Policy} \\
\toprule
 Alpha & Method & ResNet-50 & ResNet-101 & ResNet-152 &  ResNet-50  & ResNet-101 & ResNet-152 \\
\hline
0.01 & RAPS & \textbf{0.990 $\pm$ 0.001} & \textbf{0.990 $\pm$ 0.001} & \textbf{0.990 $\pm$ 0.001} & \textbf{0.990 $\pm$ 0.001} & \textbf{0.990 $\pm$ 0.001} & \textbf{0.990 $\pm$ 0.001}\\

0.01 & RAPS+TTA-Avg & \textbf{0.991 $\pm$ 0.001} & \textbf{0.990 $\pm$ 0.001} & \textbf{0.990 $\pm$ 0.001} & \textbf{0.990 $\pm$ 0.001} & \textbf{0.990 $\pm$ 0.001} & \textbf{0.990 $\pm$ 0.001}\\

0.01 & RAPS+TTA-Learned & \textbf{0.990 $\pm$ 0.001} & \textbf{0.990 $\pm$ 0.001} & \textbf{0.990 $\pm$ 0.001} & \textbf{0.990 $\pm$ 0.001} & \textbf{0.990 $\pm$ 0.001} & \textbf{0.990 $\pm$ 0.001}\\

0.05 & RAPS & \textbf{0.951 $\pm$ 0.002} & \textbf{0.952 $\pm$ 0.002} & \textbf{0.952 $\pm$ 0.002} & \textbf{0.951 $\pm$ 0.002} & \textbf{0.952 $\pm$ 0.002} & \textbf{0.952 $\pm$ 0.002}\\

0.05 & RAPS+TTA-Avg & \textbf{0.951 $\pm$ 0.001} & \textbf{0.951 $\pm$ 0.001} & \textbf{0.952 $\pm$ 0.002} & \textbf{0.951 $\pm$ 0.001} & \textbf{0.952 $\pm$ 0.002} & \textbf{0.952 $\pm$ 0.002}\\

0.05 & RAPS+TTA-Learned & \textbf{0.952 $\pm$ 0.001} & \textbf{0.952 $\pm$ 0.002} & \textbf{0.952 $\pm$ 0.002} & \textbf{0.951 $\pm$ 0.002} & \textbf{0.952 $\pm$ 0.002} & \textbf{0.952 $\pm$ 0.002}\\

0.10 & RAPS & \textbf{0.906 $\pm$ 0.004} & \textbf{0.906 $\pm$ 0.004} & 0.906 $\pm$ 0.002 & \textbf{0.906 $\pm$ 0.004} & 0.906 $\pm$ 0.004 & 0.906 $\pm$ 0.002\\

0.10 & RAPS+TTA-Avg & \textbf{0.905 $\pm$ 0.005} & 0.905 $\pm$ 0.002 & 0.908 $\pm$ 0.002 & \textbf{0.905 $\pm$ 0.004} & \textbf{0.908 $\pm$ 0.004} & \textbf{0.910 $\pm$ 0.002}\\

0.10 & RAPS+TTA-Learned & \textbf{0.905 $\pm$ 0.004} & \textbf{0.907 $\pm$ 0.003} & \textbf{0.911 $\pm$ 0.002} & \textbf{0.907 $\pm$ 0.004} & \textbf{0.908 $\pm$ 0.004} & \textbf{0.910 $\pm$ 0.002}\\

\end{tabular}
}
\caption{\textbf{Comparison of coverage across base classifiers.} Coverage values for TTA variants of conformal prediction compared to RAPS alone, across different base classifiers on ImageNet. TTA-Learned preserves coverage across all comparisons and significantly improves upon the achieved coverage using ResNet-101 with RAPS (granted, the magnitude of this improvement is small).}
\label{model-results-raps-coverage}
\end{table*}

\renewcommand{\arraystretch}{1.35}

\begin{table*}
\centering

\resizebox{.55\textwidth}{!}{%
\rowcolors{2}{gray!20}{white}

\begin{tabular}{cl|c|c|c}
\toprule
Alpha & Method & ImageNet & iNaturalist & CUB-Birds \\
\midrule
0.01 & RAPS & 0.0112 ± 0.0043 & 0.0207 ± 0.0043 & 0.0076 ± 0.0031 \\
0.01 & RAPS+TTA-Learned & 0.0113 ± 0.0067 & 0.0247 ± 0.0027 & 0.0046 ± 0.0026 \\
\hline
0.05 & RAPS & 0.2134 ± 0.0348 & 0.0609 ± 0.0217 & 0.0112 ± 0.0105 \\
0.05 & RAPS+TTA-Learned & 0.3338 ± 0.0994 & 0.0899 ± 0.0520 & 0.0350 ± 0.0412 \\
\hline
0.10 & RAPS & 0.1318 ± 0.0696 & 0.0852 ± 0.0151 & 0.2218 ± 0.1260 \\
0.10 & RAPS+TTA-Learned & 0.3198 ± 0.0977 & 0.1008 ± 0.0058 & 0.1931 ± 0.1208 \\
\bottomrule
\end{tabular}

}
\caption{\textbf{Effect of test-time augmented conformal prediction on adaptivity.} We show results in the context of ResNet-50 and RAPS, across several coverage guarantees. We compute size-stratified coverage violation (SSCV) for each run as described in Sec. \ref{sec:set-up}, and report the mean and standard deviation of SSCV across runs here. Test-time augmentation does not significantly diminish adaptivity at each coverage guarantee considered (assessed via a two-sample t-test, $p > 0.05$).}
\label{adaptivity-results}
\end{table*}

\begin{figure*}
\begin{center}
\includegraphics[width=1\textwidth]{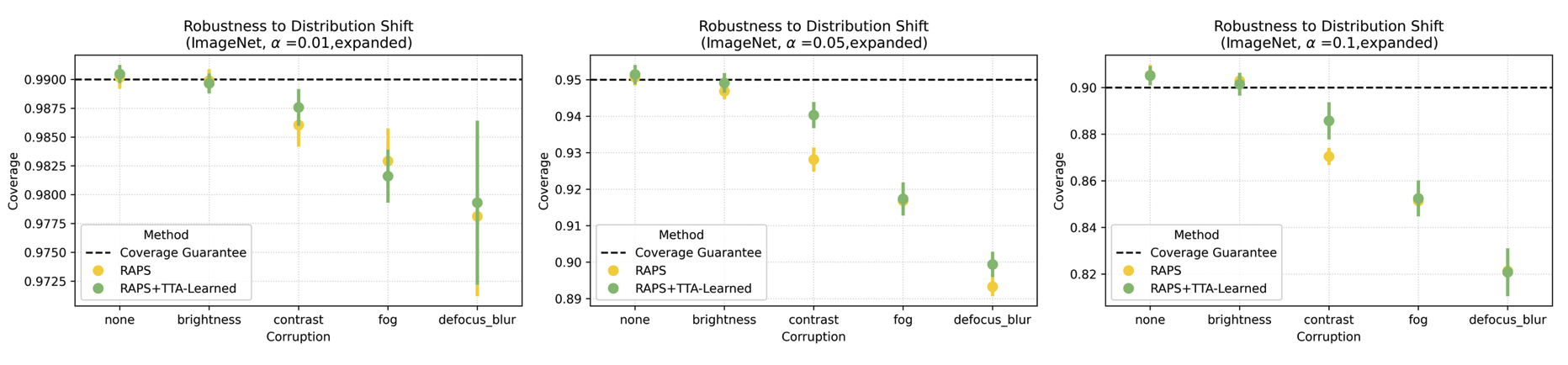}
\end{center}
\vspace{-1em}
\caption{\textbf{Impact on coverage.} We plot achieved coverage for both RAPS and RAPS+TTA-Learned across several coverage guarantees and distribution shifts. As expected, distribution shift leads conformal predictors to not meet the coverage guarantee. In each case, the addition of TTA does not worsen coverage; in some cases (for example, given the contrast corruption and a coverage guarantee of 0.05) it even improves coverage.}
\label{fig:supp_coverage_robustness}
\end{figure*}

\begin{table*}
\scriptsize
\centering
\rowcolors{2}{gray!20}{white}

\begin{tabular}{cl|c|c}
\toprule
$\alpha$ & Method & ImageNet (Expanded) & ImageNet (Simple) \\
\hline
0.01 & RAPS & 52.332 $\pm$ 8.970 & 52.332 $\pm$ 8.970\\

0.01 & RAPS+TTA-Avg & 45.604 $\pm$ 1.515 & 42.431 $\pm$ 1.516\\

0.01 & RAPS+TTA-Learned & 40.872 $\pm$ 1.377 & 40.843 $\pm$ 1.707\\

\hline
0.05 & RAPS & 8.872 $\pm$ 0.417 & 8.872 $\pm$ 0.417\\

0.05 & RAPS+TTA-Avg & 8.304 $\pm$ 0.322 & 7.945 $\pm$ 0.861\\

0.05 & RAPS+TTA-Learned & 7.723 $\pm$ 0.916 & 7.609 $\pm$ 1.027\\

\hline
0.10 & RAPS & 3.677 $\pm$ 0.104 & 3.677 $\pm$ 0.104\\

0.10 & RAPS+TTA-Avg & 3.480 $\pm$ 0.056 & 3.298 $\pm$ 0.069\\

0.10 & RAPS+TTA-Learned & 3.321 $\pm$ 0.289 & 3.348 $\pm$ 0.275\\

\bottomrule
\end{tabular}
\scriptsize
\caption{\textbf{Replicated results on MobileNetV2.} We observe trends similar to those reported to in the main text in the context of MobileNetV2. In short, RAPS combined with a learned test-time augmentation policy (RAPS+TTA-Learned) produces the smallest set sizes across the considered coverage guarantees ($\alpha \in \{0.01, 0.05, 0.10\}$) and augmentation policies.}
\label{tab:mobilenetv2}
\end{table*}

\end{document}